\documentclass[journal]{IEEEtran}

\usepackage{times}
\usepackage[numbers]{natbib}
\usepackage{multicol}
\usepackage[bookmarks=true]{hyperref}
\usepackage{graphicx}
\usepackage{amsbsy}
\usepackage{amsmath}
\usepackage{amssymb}
\usepackage{caption}
\usepackage[subrefformat=simple, labelformat=simple]{subcaption}
\usepackage{floatrow}
\usepackage{todonotes}

\DeclareFloatSeparators{space}{}

\usepackage{textcomp}
\usepackage{algorithm}
\usepackage{algpseudocode}
\usepackage{color}
\usepackage{soul}
\usepackage{xspace}

\def\eg{\emph{e.g.}}
\def\ie{\emph{i.e.}}

\newcommand{\SO}{\mathrm{SO}}
\renewcommand{\so}{\mathfrak{so}}
\newcommand{\SE}{\mathrm{SE}}
\newcommand{\se}{\mathfrak{se}}
\newcommand{\Real}{\mathbb{R}}
\newcommand{\ImageDomain}{\Omega_{s}}

\DeclareMathOperator{\Tr}{tr}

\newcommand{\Image}{\mathbf{I}}
\newcommand{\Depth}{\mathbf{D}}

\newcommand{\px}{\mathbf{u}} 
\newcommand{\pt}{\mathbf{p}} 

\newcommand{\proj}{\pi}
\newcommand{\residual}{r}
\newcommand{\residualvec}{\mathbf{r}}

\newcommand{\T}{\mathbf{T}}

\newcommand{\twist}{\boldsymbol\xi} 

\newcommand{\Tmean}{\bar{\T}} 
\newcommand{\Tes}{\hat{\T}}   
\newcommand{\TSigma}{\boldsymbol\Sigma}
\newcommand{\NormDist}{\mathcal{N}}
\newcommand{\SEJac}{\mathcal{J}}
\newcommand{\IJac}{\mathbf{J}}
\newcommand{\Information}{\boldsymbol\Lambda}

\title{\LARGE \bf
Perception-aware Path Planning 
}

\author{ \parbox{1.3 in}{\centering Gabriele Costante}
        \parbox{1.3 in}{ \centering Christian Forster}  
        \parbox{1.3 in}{ \centering Jeffrey Delmerico}  
        \parbox{1.1 in}{\centering Paolo Valigi
        \thanks{G. Costante and P. Valigi are with the Department of Engineering, University of Perugia, Italy.}}
        \parbox{1.5 in}{ \centering Davide Scaramuzza
        \thanks{C. Forster, J. Delmerico, and D. Scaramuzza are with the Robotics and Perception Group, University of Zurich, Switzerland.}}
}

\begin{document}

\maketitle


\begin{abstract}

In this paper, we give a double twist to the problem of planning under uncertainty. 
State-of-the-art planners seek to minimize the localization uncertainty by only considering the geometric structure of the scene. 
In this paper, we argue that motion planning for \emph{vision-controlled} robots should be \emph{perception aware} in that the robot should also favor texture-rich areas to minimize the localization uncertainty during a goal-reaching task. 
Thus, we describe how to optimally incorporate the \emph{photometric information} (i.e., texture) of the scene, in addition to the the geometric one, to compute the uncertainty of vision-based localization during path planning.
To avoid the caveats of feature-based localization systems (i.e., dependence on feature type and user-defined thresholds), 
we use \emph{dense, direct methods}. This allows us to compute the localization uncertainty directly from the intensity values of every pixel in the image. 
We also describe how to compute trajectories online, considering also scenarios with no prior knowledge about the map.
The proposed framework is general and can easily be adapted to different robotic platforms and scenarios. The
effectiveness of our approach is demonstrated with extensive experiments in both simulated and real-world environments using a vision-controlled micro aerial vehicle.

\end{abstract}

\IEEEpeerreviewmaketitle

\section{Introduction}

Most of the literature on robot vision has focused on the problem of \emph{passive} localization and mapping from a predefined set of view points---also known as visual odometry or SLAM 
\cite{Scaramuzza11ram}---where impressive results have been demonstrated over the last decade \cite{klein2007parallel, strasdat2010scale, forster2014svo, engel2014lsd}.
Minor work has instead tackled the problem of how to \emph{actively} control the perception pipeline in order to improve the performance of a given task \cite{bajcsy1988active,Soatto11active}.

\begin{figure}[tb]
  \centering
  \ffigbox{}
  {
    \CommonHeightRow
    {
      \begin{subfloatrow}[3]
        \ffigbox[\FBwidth]
        {\includegraphics[height=\CommonHeight]{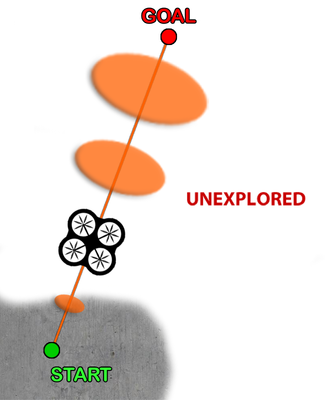}}
        {\caption{}\label{fig:overview_init}}
        \ffigbox[\FBwidth]
        {\includegraphics[height=\CommonHeight]{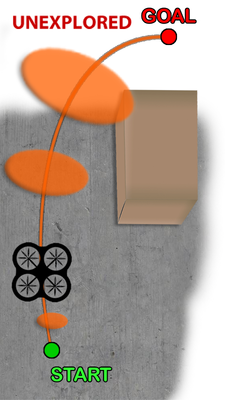}}
        {\caption{}\label{fig:overview_explore}}
        \ffigbox[\FBwidth]
        {\includegraphics[height=\CommonHeight]{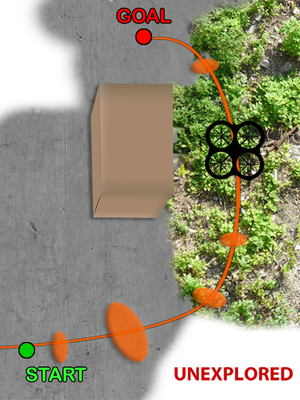}}
        {\caption{}\label{fig:overview_texture}}

      \end{subfloatrow}
    }
    \caption{\small Online perception-aware path planning: An initial plan is
      computed without prior knowledge about the environment \subref{fig:overview_init}. 
      The plan is then updated as new obstacles \subref{fig:overview_explore} or new textured 
      areas \subref{fig:overview_texture} are discovered. Although the new trajectory (c) is longer 
      than the one in (b), it contains more \emph{photometric} information and, thus, is optimal with 
      respect to the pose localization uncertainty. 
}
 \label{main_idea}
  }
\end{figure}

In this paper, we address the problem of how to optimally leverage vision in a goal-reaching task to select trajectories with minimum localization accuracy.
State-of-the-art path planners seek to minimize the localization uncertainty by only considering the geometric structure of the scene. 
However, for vision-controlled robots it is crucial to also consider the photometric appearance (i.e., texture) of the environment when designing reliable trajectories (cf. Figure \ref{main_idea}).

\begin{figure}[t]
  \centering
  \ffigbox{}
  {
    \CommonHeightRow
    {
      \begin{subfloatrow}[2]
        \ffigbox[\FBwidth]
        {\includegraphics[height=\CommonHeight]{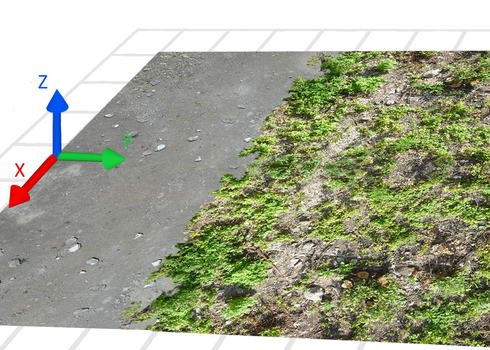}}
        {\caption{}\label{fig:overview_inf_gain_a}}
        \ffigbox[\FBwidth]
        {\includegraphics[height=\CommonHeight]{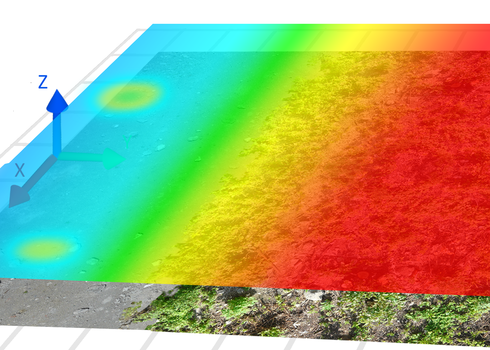}}
        {\caption{} \label{fig:overview_inf_gain_b}}

      \end{subfloatrow}
    }
    \caption{\small \subref{fig:overview_inf_gain_a} A scene and \subref{fig:overview_inf_gain_b} its localization uncertainty (notably, the trace of the covariance matrix) for a downward-looking camera at a given height.
		The localization uncertainty is visualized as a heat-map (blue means high uncertainty, red means low).}
    \label{overview_inf_gain}
  }
\end{figure}

The basic observation is that the uncertainty of vision-based localization is strongly affected by the photometric appearance of the observed scene (cf. Figure \ref{overview_inf_gain}).
Thus, highly-textured areas should be preferred to locations with poor photometric information when planning reliable trajectories (i.e., with low localization uncertainty).
Driven by this observation, we aim to answer the following question: \emph{What is the trajectory that minimizes the camera pose-estimation uncertainty in a robot-navigation task?}
In practice, the \emph{best} trajectory depends on different factors: 
(i) the current robot pose and uncertainty, (ii) the geometry of the scene, and (iii) the photometric appearance of the scene.
Based on the these considerations, we describe how to incorporate the photometric information, in addition to the the geometric one, to compute the uncertainty of vision-based 
localization during path planning.
The best trajectory can then be computed as a function of the robot's current pose and the expected pose-uncertainty reduction due to the predicted 3D structure and photometric 
appearance of the scene (see Figure \ref{main_idea}).

Since we want to handle scenarios with no prior knowledge about the map, we also present an online adaptation of the proposed framework. 
In particular, we update the plan as the robot explores the scene, adapting the perception-aware trajectory as new photometric information becomes available.

\section{Related work}
\label{sec:related}

 \subsection{Planning in Information Space}
The selection of trajectories that minimize the localization uncertainty is often referred to as ``Planning under Uncertainty'' or ``Planning in Information Space''.
This problem has generally been solved with Partially Observable Markov Decision Processes (POMDPs) or 
through graph-search in the belief space \cite{Bonet00planningwith}.
While these approaches are well-established, in general their computational complexity grows exponentially in the number of possible actions and observations.
To overcome this issues, Rapidly-exploring Random Tree (RRT*) \cite{karaman2011sampling} were introduced to perform fast trajectory computation and guarantee asymptotic optimality.
Furthermore, Rapidly-exploring Random Belief Trees (RRBTs) were proposed by \cite{bry2011rapidly} as an extension of the RRT* framework to take into account the pose uncertainty.
However, while the RRBTs are well-suited for energy minimization tasks,
in this work, we specifically focus on selecting trajectories that maximize the visual information without considering robot dynamics or control efforts.
Thus, we choose to extend the RRT* framework to take into account also the pose uncertainty when computing optimal trajectories.

\subsection{Active Perception}
When perception is incorporated into the path planning process, the problem of selecting optimal viewpoints to maximize the performance of a given task is referred to as 
\emph{active perception} \cite{bajcsy1988active, blake1993active, aloimonos1988active, chen2011active, soatto2013actionable}.
One of the goals of active perception is \emph{active localization}, which seeks to compute control actions and trajectories that minimize the pose estimation uncertainty.
Most active localization works have been in the context of robot SLAM or exploration.
Depending on the sensor used, they can be classified into range-based \cite{feder1999adaptive,bourgault2002information,bachrach2012estimation} 
or vision-based \cite{davison2002simultaneous,vidalcalleja2010,mostegelactive,sadat2014feature,achtelik2014motion}.

While range sensors only perceive the geometric structure of the environment, vision sensors are more informative because they can capture both the geometry and appearance of a scene.
Davison and Murray \cite{davison2002simultaneous} were the first to take into account the effects of actions during visual SLAM. The goal was to select a
fixation-point of a moving stereo head attached to a mobile robot in order to minimize the motion drift along a predefined trajectory.
Vidal Calleja et al. \cite{vidalcalleja2010} demonstrated an active feature-based visual SLAM framework that provides realtime user-feedback to minimize both map and camera pose uncertainty.
Bryson and Sukkarieh \cite{Bryson08taes} demonstrated a similar visual and inertial EKF-SLAM formulation for active control of flying vehicles. The goal was to cover a predefined area with a camera while maintaining an accurate estimation of
both the map and the vehicle state. Extensive simulation results were provided of a MAV that is restricted to fly on a plane.
Mostegel et al. \cite{mostegelactive} proposed a set of criteria to estimate the influence of camera motion on the stability of visual localization for MAVs.

The minimization of the pose covariance in vision-based path-planning systems was addressed in \cite{sadat2014feature} and \cite{achtelik2014motion}.
Achtelik et al. \cite{achtelik2014motion} used RRBTs to evaluate offline multiple path hypotheses in a known map and select paths with minimum pose uncertainty while at the same time considering the vehicle dynamics. They computed the pose covariance directly from bundle adjustment, by minimizing the reprojection errors of the 3D map points across all images.
The approach was demonstrated on a MAV.
Sadat et al. \cite{sadat2014feature} proposed a strategy to plan trajectories for MAVs, which prefers paths rich of visual features.
A viewpoint score based on the number of observed features was used to measure the quality of localization. The system used RRT* to iteratively re-plan as the robot explored the environment. 
As a fixed part of the previous plan is executed, RRT* is recomputed from scratch.


\subsection{Feature-based vs Dense, Direct Methods}
All vision-based works previously mentioned represent the scene as a set of \emph{sparse} 3D landmarks corresponding to discriminative features in the observed images (e.g., SIFT, SURF, etc.)
and estimate structure and motion through reprojection-error minimization.
A reason for the success of these methods is the availability of robust feature detectors and descriptors that allow matching images with large disparity. 
The disadvantage of feature-based approaches is the dependence on the feature type, the reliance on numerous detection and matching thresholds, the necessity for robust estimation techniques to deal with incorrect correspondences (e.g., RANSAC),
and the fact that most feature detectors are optimized for speed rather than precision.

The alternative to feature-based methods is to use \emph{dense, direct methods} \cite{irani1999direct}.
Direct methods have the advantage that they estimate structure and motion directly by minimizing an error measure (called \emph{photometric error}) that is
based on image’s pixel-level intensities. The local intensity gradient magnitude and direction is used in the optimization
compared to feature-based methods that only consider the distance to a feature-location. 
Pixel correspondence is given directly by the geometry of the problem, eliminating the need for robust data association techniques.
Direct methods are said \emph{dense} if they exploit the visual information even from areas where gradients are small (i.e., not just edges). 
Dense, direct methods have been shown to outperform feature-based methods in terms of robustness in scenes with little texture \cite{lovegrove2011}
or in the case of camera defocus and motion blur \cite{newcombe2011dtam,Meilland13iccv}.
Using dense, direct methods, the 6-DoF pose of a camera can be recovered by \emph{dense image-to-model alignment}, 
which is the process of aligning the observed image to a view synthesized from the estimated 3D map through photometric error minimization.


The first approach taking advantage of dense, direct methods in the context of active perception was proposed by Forster et al. \cite{forsterActive2014}. 
However, the task was specified in terms of maximizing the quality of the map (i.e., minimizing the map uncertainty). Thus, the robot localization uncertainty was not considered. 
Additionally, path planning from a start to a goal point was not investigated.
Conversely, in this paper we are interested in computing trajectories towards a predefined goal while minimizing the robot pose uncertainty along the path.
In contrast to previous works based on sparse features, we use dense, direct methods.

\subsection{Contributions}

Our contributions are:
\begin{itemize}
 \item An \emph{online perception-aware path planning} framework that computes the best path towards a predefined goal
through the exploitation of both the geometric \emph{and photometric} information (i.e., texture) of the scene. 
 To the best of our knowledge, this is the first attempt to use the photometric appearance in addition to the geometric 3D structure for planning under uncertainty.
 \item We use \emph{dense, direct methods} to compute the photometric information gain directly from the intensity values of every pixel in the image. This avoids the caveats of feature-based localization systems, such as the dependence on the type of feature detector and descriptor and the reliance on user-defined thresholds for detection and matching.
 \item We integrate the Lie Group-based propagation proposed in {\cite{barfoot_tro14}} and we extend the Rapidly-exploring Random Tree (RRT*) {\cite{karaman2011sampling}} framework to take into account the pose uncertainty when computing trajectories.
\item We implement and demonstrate the effectiveness of our approach on an actual vision-based quadrotor performing vision-based localization, dense 3D reconstruction, and online perception-aware planning.
\end{itemize}

This paper extends the work presented in \cite{costante2015isrr} with more experimental results and technical details.

\subsection{Outline}
The outline of the paper is as follows: in Section \ref{lie_group_sec}, we introduce the Lie-Group--based propagation framework and describe how the pose uncertainties are propagated along the trajectory.
Section \ref{information_gain} describes the dense image-to-model alignment strategy to compute the photometric information gain in terms of the scene texture.
In Section \ref{planning_sec}, we adapt the RRT* framework to generate trajectories that minimize the camera pose uncertainty given the photometric information computed along the path.
In Section \ref{experiments_sec}, we present the experimental evaluation. Finally, in Section \ref{conclusion_sec} we draw conclusions and highlight possible future improvements.

\section{Lie group based Uncertainty Propagation}\label{lie_group_sec}

Different trajectories lead to different evolutions of pose covariance.
For this reason, it is crucial to predict how the pose uncertainty will be affected given a candidate route.
To achieve this, we need a state representation to propagate the pose estimate, together with its uncertainty, when executing a predefined trajectory. 
  
When choosing a state representation, most challenges arise because the rotation parametrizations have either singularities or constraints. 
This is related to the fact that rotation variables are not vectors but members of a non-commutative group, \ie,  the Lie group $\SO(3)$. 
As a consequence, using a first-order approximation to propagate the covariance matrix (\eg, in standard EKFs) does not guarantee a good estimate of the uncertainty.
Conversely, Monte Carlo techniques are more reliable, but the computational effort required to reach a realistic estimate is often unacceptable.
We can achieve both a robust and an efficient representation if we preserve the nature of the rotation matrices, \ie, we represent the robot poses as Lie group members.


\subsection{Associating Uncertainty to Rigid Body Motions}

First of all, we provide some assumptions and preliminary notations that we use in our formulations in the following sections.

We represent the pose of the robot as a 6 Degree of Freedom (DoF) transformation matrix $\T$, member of the \emph{special Euclidean group} in $\Real^3$, which is defined as follows:

\begin{equation}
  \begin{aligned}
    \SE(3) := \left \{ \mathbf{T} = \left [\begin{array}{cc}
					      \mathbf{C} & \mathbf{r} \\
					      \mathbf{0}^T & 1
					      \end{array} \right] \ \middle| \ {\mathbf{C} \in \SO(3), \mathbf{r}}  \in  \mathbb{R}^3
					      \right \},
  \end{aligned}
\end{equation}

where 

\begin{equation}
  \begin{aligned}
    \SO(3) :=\big\{ \mathbf{C} \in \mathbb{R}^{3 \times 3} \ \big| \ \mathbf{C} \mathbf{C}^T = \mathbf{1}, \det \mathbf{C} = 1 \big\}
  \end{aligned}
\end{equation}

is the special orthogonal group in $\Real^3$ (the set of spatial rotations) and $\mathbf{1}$ is the $3 \times 3$ identity matrix.

In the following, the \emph{Lie Algebra} associated to the $\SE(3)$ Lie Group is referred as $\se(3)$.
To represent the uncertainty of the robot pose, we use the formulation proposed in \cite{barfoot_tro14}.
We define a random variable for $\SE(3)$ members according to:

\begin{equation}
  \begin{aligned}
    \mathbf{T} := \exp(\twist^\wedge)\mathbf{\bar{T}}
  \end{aligned}
\end{equation}
In this definition, $\mathbf{\bar{T}}$ is a noise-free value that represents the mean of the pose, while $\twist \in  \mathbb{R}^6$ is a small perturbation in the tangent space that we assume to be normally 
distributed with zero mean and covariance $\TSigma$.
We make use of the $\wedge$ operator to map $\twist$ to a member of the Lie algebra $\se(3)$ using:

\begin{equation}
  \begin{aligned}
    \twist^{\wedge} 
    := \left[\begin{array}{c} \boldsymbol\rho  \\ \boldsymbol\phi \end{array} \right]
    =  \left[\begin{array}{cc}
					   \boldsymbol\phi^{\wedge} & \boldsymbol\rho  \\
					   \mathbf{0}^T & 0
					   \end{array} \right] \mbox{,}
  \end{aligned}
\end{equation}

where $\boldsymbol\phi$ is a member of the \emph{Lie algebra} $\so(3)$:

\begin{equation}
 \begin{aligned}
  {\boldsymbol\phi}^\wedge := \left [ \begin{array}{c}
				      \phi_1  \\
				      \phi_2 \\
				      \phi_3
				      \end{array} \right]^{\wedge}  = \left [\begin{array}{ccc}
				      0 & -\phi_3 & \phi_2  \\
				      \phi_3 & 0 & -\phi_1 \\
				      -\phi_2 & \phi_1 & 0
				      \end{array} \right]
  \end{aligned}
\end{equation}
Observe that the operator $\wedge$ is 'overloaded' and can be applied to both $6 \times 1$ and $3 \times 1$ vectors \cite{barfoot_tro14, chirikjian2011stochastic}. They are disambiguated by the context. 

Furthermore, we indicate with $\T_{k,w}$ the robot pose at time $k$ relative to the world frame $w$ and with $\T_{k+1,k}$ the transformation between the pose at time $k$ and $k+1$. 


\subsection{Pose Propagation}\label{lie_group_propagation}

\begin{figure}[t]
  \centering
  \ffigbox{}
  {
    \CommonHeightRow
    {
      \begin{subfloatrow}[2]
        \ffigbox[\FBwidth]
        {\includegraphics[height=\CommonHeight]{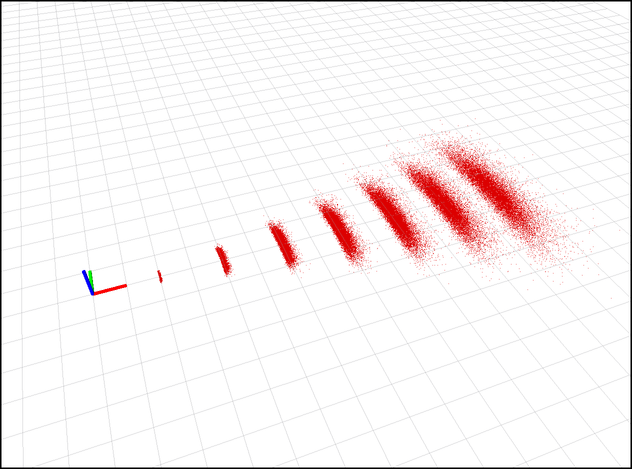}}
        {\caption{}\label{fig:propagation_example_2d}}
        \ffigbox[\FBwidth]
        {\includegraphics[height=\CommonHeight]{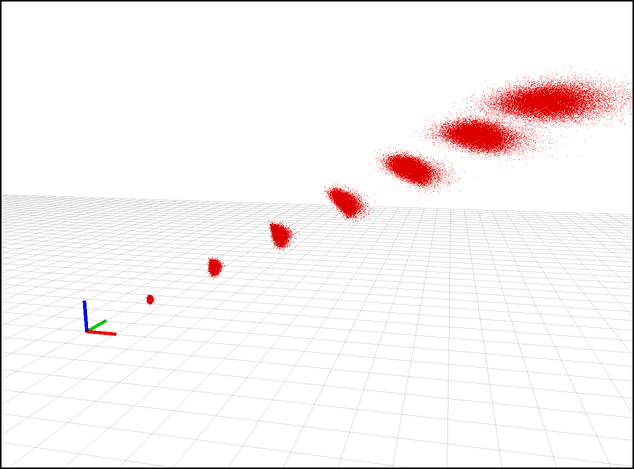}}
        {\caption{} \label{fig:propagation_example_3d}}

      \end{subfloatrow}
    }

    \caption{\small Examples of propagation using the fourth-order Lie group framework. The two columns show two different propagation tests.
      In \ref{fig:propagation_example_2d}, the covariance is propagated after 100 motions of 1 meter along the $x$ axis, with a motion uncertainty of $\TSigma_{k+1,k} =  \mathbb{\mathbf{diag}}(0,0,0,0,0,0.03)$. 
      In \ref{fig:propagation_example_3d}, we perform 100 motions ($(1.0, 0.0, 0.1)$ meters) starting from the pose $(0,0,0,0,0,\pi/8)$, and with $\TSigma_{k+1,k} = \mathbb{\mathbf{diag}}(0.01,0.01,0.01,0.001,0.001,0.03)$.
      The covariances are depicted as point clouds, sampling the distributions every 10 motions. \vspace{-1.5em}
}
    \label{propagation_example}
  }
\end{figure}

Properly modeling the uncertainty propagation according to the IMU odometry model would require the extension of the robot state vector with the instantaneous velocity.
However, to reduce the problem complexity, we assume in the following that the velocity remains constant and, thus, 
the odometry uncertainty, denoted by $\TSigma_{k+1,k}$, associated to all motions $\T_{k+1,k}$, is fixed.
 
Given the transformation $\T_{k+1,k}$, we reason about the propagation of the mean and the covariance of the resulting pose $\T_{k+1,w}$.
Assuming no correlation between the current pose and the transformation between $k$ and $k+1$, we can consider $\T_{k,w}$ and $\T_{k+1,k}$ as represented by their means and covariances:
\begin{equation}
  \{ \Tmean_{k,w}, \TSigma_{k,w} \}, \ \{ \Tmean_{k+1,k}, \TSigma_{k+1,k} \}.
\end{equation}
Combining them, we get
\begin{equation}
  \T_{k+1,w} = \T_{k,w} \ \T_{k+1,k}.
\end{equation}

To compute the mean and the covariance of the compound pose, we use the results from \cite{barfoot_tro14}.
The mean is
\begin{equation}
  \Tmean_{k+1,w} = \Tmean_{k,w} \ \Tmean_{k+1,k}, \label{mean_prop}
\end{equation}
and the covariance, approximated to fourth order, is
\begin{equation}
  \TSigma_{k+1,w} \backsimeq \TSigma_{k,w} + \mathcal{T} \TSigma_{k+1,k}\mathcal{T}^\top + \mathcal{F} \label{covariance_propagation}
\end{equation}
where $\mathcal{T}$ is $Ad(\Tmean_{k,w})$, \ie, the adjoint operator for $\SE(3)$, and $\mathcal{F}$ encodes the fourth-order terms.  
Equations (\ref{mean_prop}) and (\ref{covariance_propagation}), we can propagate the uncertainty along a nominal trajectory. 
Figure \ref{propagation_example} depicts examples of covariance propagations.



\subsection{Measurement Update} \label{information_gain}

In this section, we describe the computation of the photometric information associated to a measurement at a particular viewpoint in order to update the predicted pose uncertainty.
The measurement process defines the information that can be obtained from images, hence, we summarize it in the following.
In contrast to previous works based on \emph{sparse} keypoints, we use a dense image-to-model alignment approach for the measurement update, which uses the intensity and depth of \emph{every} pixel in the image.

\subsubsection{Preliminary Notation}

At each iteration of the navigation process,
we can compute a dense surface model $\mathcal{S} \in \Real^3 \times \Real^{+}$ (3D position and grayscale intensity) relative to the explored part of the scene (see Figure \ref{fig:inf_gain_a}).
The rendered synthetic image is denoted with $\Image_s : \ImageDomain \subset \Real^2 \rightarrow \Real^{+}$, where $\ImageDomain$ is the image domain and $\px = (u,v)^T \in \ImageDomain$ are pixel coordinates.
Furthermore, we refer to the depthmap $\Depth_s$, associated to an image $\Image_s$, as the matrix containing the distance at every pixel to the surface of the scene:
\begin{equation}
   \Depth_s :\ImageDomain \rightarrow \Real^{+} \text{; } \px \mapsto  d_{\px},  \label{depth_def}
\end{equation}
where $d_{\px}$ is the depth associated to $\px$.
Note that, since we need to predict the uncertainty propagation during the planning phase, the actual image at a given location is not available at the beginning. 
As a consequence, we synthesize the predicted image for each waypoint selected using the reconstructed map and we update the pose uncertainty estimates accordingly.

A 3D point $\pt$~$=$~$(x,y,z)^T$ in the camera reference frame is mapped to the corresponding pixel in the image $\px$ through the 
camera projection model $\proj : \Real^3 \rightarrow \Real^2$
\begin{equation}
   \px = \proj(\pt).  \label{proj_equation}
\end{equation}
On the other hand, we can recover the 3D point associated to the pixel $\px$ using the inverse projection function $\proj^{-1}$ and the depth $d_{\px}$:
\begin{equation}
   \pt_{\px} = \proj^{-1}(\px, d_{\px}). \label{inv_proj_equation}
\end{equation}
Note that the projection function $\proj$ is determined by the intrinsic camera parameters that are known from calibration.

Finally, a rigid body transformation $\T \in \SE(3)$ rotates and translates a point $\mathbf{q}$ as follows:
\begin{equation}
  \mathbf{q}'(\T) := (\mathbf{1}\, |\, \mathbf{0})\,\T\,(\mathbf{q}^T ,1)^T.  \label{transform_point}
\end{equation}


\subsubsection{Dense Image-to-Model Alignment}

\begin{figure}[t]
  \begin{center}
    \includegraphics[width=0.59\columnwidth]{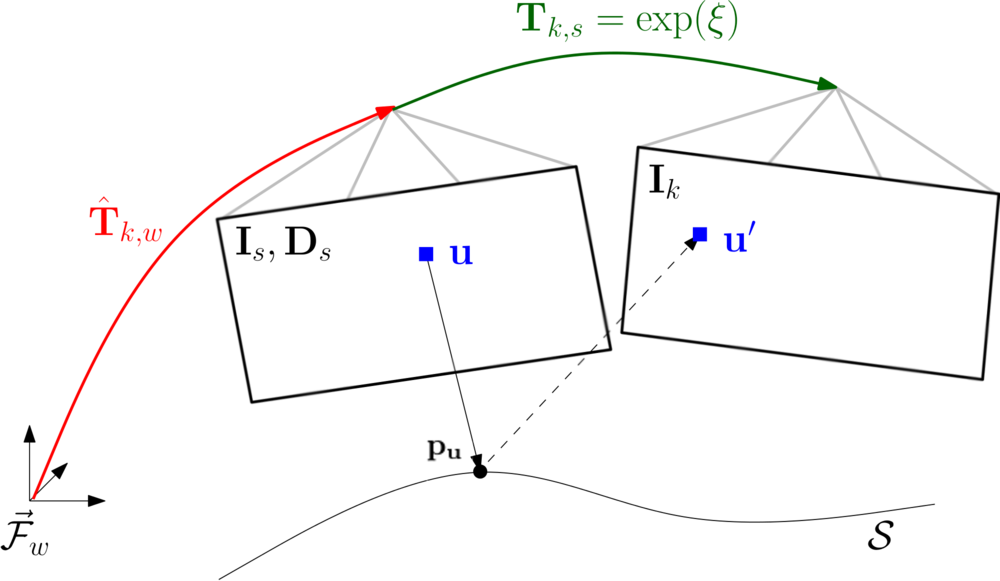}
    \caption{\small Illustration of the dense image-to-model alignment used in the measurement update. Given an estimate of the pose $\Tes_{k,w}$, 
    we can synthesize an image and depthmap $\{\Image_k,\mathbf{D}_k\}$ from the 3D model $\mathcal{S}$. 
    The best update $\twist$ of the pose estimate is computed by minimizing the intensity difference of corresponding pixels $\{\px,\px'\}$.}
    \label{fig:img_align}
  \end{center}
\end{figure}

Given the dense 3D model of the environment we can synthesize an image and the relative depthmap $\Image_s $, $\Depth_s$ at the estimated pose of the camera $\T_{k,w}$.
To refine the current pose estimate $\Tes_{k,w}$, of the frame $k$ with respect to the global world frame $w$, 
we use dense image-to-model alignment \cite{newcombe2011dtam, meilland2013unifying} (see Figure~\ref{fig:img_align}). 
This approach determines the incremental updates $\twist$ to the current pose estimate by minimizing the photometric error between the observed image and the synthetic one.
Once converged, this approach also provides the uncertainty of the alignment through evaluation of the \emph{Fisher Information Matrix}, which is used in our approach to select informative trajectories.
The image residual $\residual_\px$ for a pixel $\px$ is the difference of the intensity value at pixel $\px$ in the real image acquired 
at time step $k$ and the intensity value in the synthetic image rendered at the estimated position $\Tes_{k,w}$:
\begin{equation}
  \residual_{\px} = \Image_k(\px) - \Image_s(\proj(\pt'_\px(\Tes_{k,w}))) \label{img_res}
\end{equation}
The residual is assumed to be normally distributed $\residual_{\px}$~$\sim$~$\NormDist(0,\sigma_i^2)$, where $\sigma_i$ is the standard deviation of the image noise.

\begin{figure}[t]
  \centering
  \ffigbox{}
  {
    \CommonHeightRow
    {
      \begin{subfloatrow}[2]
        \ffigbox[\FBwidth]
        {\includegraphics[height=\CommonHeight]{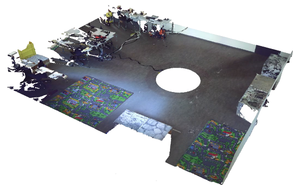}}
        {\caption{}\label{fig:inf_gain_a}}
        \ffigbox[\FBwidth]
        {\includegraphics[height=\CommonHeight]{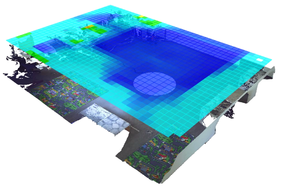}}
        {\caption{} \label{fig:inf_gain_b}}

      \end{subfloatrow}
    }
    \caption{\small Figure \ref{fig:inf_gain_b} shows the information gain related to the scene in \ref{fig:inf_gain_a} (Figure \ref{fig:real_exp_setup}.a) in the case of fixed height.}
    \label{fig:inf_gain}
  }
\end{figure}

The dense image-to-model alignment approach computes the pose $\T_{k,w}$ of the synthetic image $\Image_s$, which minimizes the residual with the actual image and, hence, the pose of the robot.
Due to the nonlinearity of the problem, we assume that we have an initial guess of the pose $\Tes_{k,w}$ and iteratively compute update steps $\twist^\wedge\in\se(3)$
\begin{equation}
  \Tes_{k,w} \leftarrow \exp(\twist^\wedge)\Tes_{k,w}
\end{equation}
that minimize the residual.
The update step minimizes the following least-squares problem
\begin{equation}\label{eq:leastsquares}
    \twist = \arg\min_{\twist} \sum_{\px \in \ImageDomain} 
    \frac{1}{2\sigma_i^2}\Big[\Image_k(\px') - \Image_s(\proj(\pt'_\px(\Tes_{k,w}))) \Big]^2 
\end{equation}
with $\pt_{\px}$ given by \eqref{inv_proj_equation}, $\pt_{\px}'$ as in \eqref{transform_point}, and
\begin{equation}
     \px' = \proj\bigl(\pt'_{\px}(\exp(\twist^\wedge))\bigr).
 \end{equation}

Addressing the least-squares problem \eqref{eq:leastsquares} using the Gauss-Newton method leads to the normal equations that can be solved for $\twist$:
\begin{equation}
  \IJac^T\IJac \twist = -\IJac^T \residualvec,
\end{equation}
where $\IJac$ and $\residualvec$ are the stacked Jacobian and image residuals of all pixels $\px \in \ImageDomain$ respectively.

Specifically, the least-squares minimization requires the computation of the Jacobian of the residual in~\eqref{eq:leastsquares} at each pixel $\px$, which 
can be written as a function of the gradient in the observed image and the synthetic depthmap\footnote{see Appendix B in \cite{strasdatPhD} for a detailed derivation of the exponential map Jacobian computation.}:
\begin{equation}\label{eq:measurement_jacobian}
  \IJac_\px
  = \bigl(\nabla\Image_k(\px)\bigl)^{T} 
  \frac{\partial\proj(\mathbf{b})}{\partial \mathbf{b}}\bigg|_{\mathbf{b}=\pt_{\px}'}
   \frac{\partial\pt'_{\px}\bigl(\exp(\twist^\wedge)\bigr)}{{\partial\twist}}\bigg|_{\twist=\mathbf{0}}
\end{equation}
In this work, for sake of simplicity, we assume depth uncertainty to be zero. However, non-zero values can easily be integrated into our framework.

At the convergence of the optimization, the quantity
\begin{equation}
   \Information_k = \frac{1}{\sigma_i^2}\IJac^T\IJac \label{ing_equation}
\end{equation}
is the \emph{Fisher Information Matrix} \cite{fisher1922mathematical} and its inverse is the covariance matrix $\TSigma_{\Image_k}$ of the measurement update.

According to \cite{barfoot_tro14}, we find the covariance matrix after the measurement update at time $k$ by computing 
\begin{equation} \label{eq:cov_incorp}
   \TSigma_{k,w} \leftarrow \Big( \Information_k^{-1} + \SEJac^{-T} \TSigma_{k,w} \SEJac^{-1} \Big)^{-1},
\end{equation}
where the ``left-Jacobian $\SEJac$ is a function of how much the measurement update modified the estimate.
Note that the information is not only a function of the image gradient but also of the depth at every
pixel (see last term in \eqref{eq:measurement_jacobian}). However, the uncertainty in the orientation is only a function of the texture and independent of the depth.

Solving the dense image-to-model alignment optimization, allows us to estimate the camera pose during execution of the trajectory, by
means of iteratively synthesizing synthetic images from the environment model, and to refine the alignment.
However, during planning, the location of viewpoints evaluated along a trajectory is known and only the computation of the uncertainty in \eqref{eq:cov_incorp} is relevant.
Therefore, the photometric information $\Information_k$ can directly be incorporated into the pose covariance with Equation \eqref{eq:cov_incorp}.

Given the information matrix in \eqref{eq:cov_incorp}, we define the photometric information gain as $\Tr(\Information_k)$.
Figure \ref{fig:inf_gain_b} depict the photometric information gain map for the scenario in Figure \ref{fig:inf_gain_a}.
  \section{Planning under Uncertainty}\label{planning_sec}
  
  Thanks to the propagation framework described in the previous sections, we are able to predict the pose uncertainty after sequences of camera motions.
  Furthermore, we can update the pose covariance according to the expected photometric information gain computed with the dense image-to-model alignment strategy presented in Section \ref{information_gain}. 
  To compute the optimal path we need to evaluate all possible trajectories and we need to do that \emph{efficiently}. 
  In the following, we describe how the sequence of viewpoints that minimize the localization uncertainty is selected with low complexity.
  Furthermore, as we do not assume to have any given prior knowledge about the scene, the photometric information of the environment, as well as its 3D geometry, are unknown.
  Hence, the trajectory that is considered optimal in the beginning will be adapted as new information is gathered by the robot.
  
  
  As stated in the previous sections, RRT* provides an efficient framework to efficiently compute trajectories.
  Nevertheless, in its original formulation, the RRT* does not take into account the pose uncertainty.
  
  To benefit from the RRT* advantages and overcome its limitations, we adapt this framework in the next section to our scenario, proposing a cost function that encodes
  both the distance term and the amount of uncertainty associated with a candidate path. 
  
  \subsection{Perception-aware RRT*}
  \label{perception-rrbt}
  
   \begin{algorithm}[t]
  \caption{\enspace Perception-aware RRT*}
\begin{small}
 \begin{algorithmic}
  \State
01: \textbf{Init:} Initial vertex $v_0.x = x_{\text{init}}$; $v_0.p = \text{root}$; \\
\ \ \ \ \ Initial pose covariance $v_0.\Sigma = \Sigma_0$; Initial cost $v_0.c = 0$;  \\
\ \ \ \ \ Initial Vertex set $V = \{v_0\}$; Number of iterations $T$; \\
\ \ \ \ \ Collision radius $c$\State
02: \textbf{for} $t=1,\ldots, T$ \textbf{do} \State
03: \ \ $x_{\text{new}} = \text{Sample}()$ \State
04: \ \ $v_{\text{nst}} = \text{Nearest}(x_{\text{new}})$ \State
05: \ \ \textbf{if} $\text{ObstacleFree}(v_{\text{new}}$, $v_{\text{nst}}$, $c$) \State 
06: \ \ \ \ $\Sigma_{\text{t}} = \text{Propagate}(v_{\text{nst}}.x, v_{\text{nst}}.\Sigma, v_{\text{new}}.x)$ \State
07: \ \ \ \ $\Sigma_{\text{t}} = \text{Update}(\Sigma_{\text{t}}, v_{\text{new}}.\Information)$ \State
08: \ \ \ \ $J_{\text{min}} = v_{\text{nst}}.c + (1-\alpha)\Tr(\Sigma_{\text{t}}) + \alpha \text{Dist}(v_{\text{nst}}.x, v_{\text{new}}.x)$ \State
09: \ \ \ \ $v_{\text{min}} = v_{\text{nst}}$ \State
10: \ \ \ \ $V = V \cup v(x_{\text{new}})$ \State
11: \ \ \ \ $V_{\text{neighbors}} = \text{Near}(V, v_{\text{new}})$ \State
12: \ \ \ \ \textbf{for all} $v_{\text{near}} \in V_{\text{neighbors}} \textbf{do}$ \State
13: \ \ \ \ \ \ \textbf{if} $\text{CollisionFree}(v_{\text{near}}$, $v_{\text{new}}$, $c$) \State
14: \ \ \ \ \ \ \ \ $\Sigma_{\text{t}} = \text{Propagate}(v_{\text{near}}.x, v_{\text{near}}.\Sigma, v_{\text{new}}.x)$ \State
15: \ \ \ \ \ \ \ \ $\Sigma_{\text{t}} = \text{Update}(\Sigma_{\text{t}}, v_{\text{new}}.\Information)$ \State
16: \ \ \ \ \ \ \ \ \textbf{if} $v_{\text{near}}.c + (1-\alpha)\Tr(\Sigma_{\text{t}})  $\\ \ \ \ \ \ \ \ \ \ \ \ \ \ \ \ \ \ \ \ \ \ \ \ $ + \alpha \text{Dist}(v_{\text{near}}.x, v_{\text{new}}.x) < J_{\text{min}}$ \State
17: \ \ \ \ \ \ \ \ \ \ $J_{\text{min}} = v_{\text{near}}.c + (1-\alpha)\Tr(\Sigma_{\text{t}}) + \alpha \text{Dist}(v_{\text{near}}.x, v_{\text{new}}.x)$ \State
18: \ \ \ \ \ \ \ \ \ \ $v_{\text{new}}.\Sigma = \Sigma_{\text{t}} $ \State
19: \ \ \ \ \ \ \ \ \ \ $v_{\text{new}}.c = J_{\text{min}}$ \State
20: \ \ \ \ \ \ \ \ \ \ $v_{\text{min}} = v_{\text{near}} $ \State
21: \ \ \ \ \ \ \ \ \textbf{end if} \State
22: \ \ \ \ \ \ \textbf{end if} \State
23: \ \ \ \ \ \ $\text{ConnectVertices}(v_{\text{min}}, v_{\text{new}}$) \State
24: \ \ \ \ \textbf{end for} \State
25: \ \ \ \ $\text{RewireTree}()$ \State
26: \ \ \textbf{end if} \State
27: \textbf{end for} \State
 \end{algorithmic}
\end{small}
 \label{rrt}
 \end{algorithm}
  
      \begin{figure}[t]
\begin{center}
  \includegraphics[width=0.68\columnwidth]{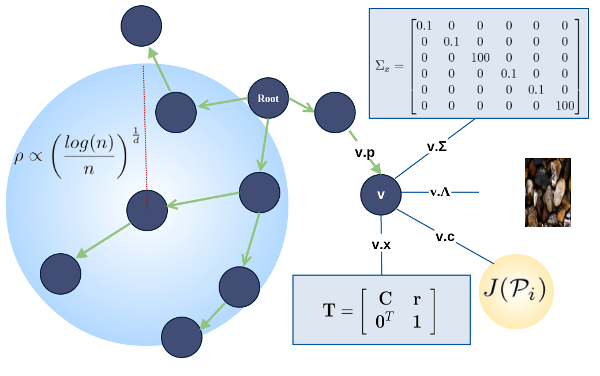}  
    \end{center}
      \caption{\small Example of a tree configuration. The green arrows connect different vertices in the tree. Each leaf has a unique path to the root. 
      The blue circle includes all the vertexes affected by the rewire procedure when a new element is sampled and added in the tree. 
      The vertex $v$ is expanded to show the properties of each node. \vspace{-1.5em}}
  \label{fig:tree_representation}
\end{figure}
  
  \begin{figure*}[th!]
  \centering
  \ffigbox{}
  {
    \CommonHeightRow
    {
      \begin{subfloatrow}[3]
        \ffigbox[\FBwidth]
        {\includegraphics[height=\CommonHeight]{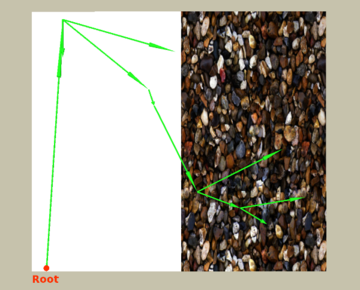}}
        {\caption{\enspace \small Standard RRT* - 10 steps} \label{fig:tree_construction_a}}
        \ffigbox[\FBwidth]
        {\includegraphics[height=\CommonHeight]{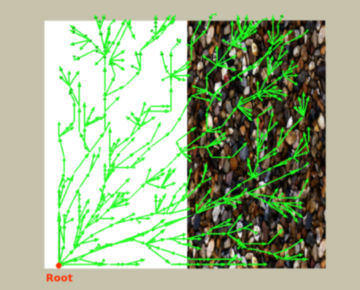}}
        {\caption{\enspace \small Standard RRT* - 500 steps} \label{fig:tree_construction_b}}
        \ffigbox[\FBwidth]
        {\includegraphics[height=\CommonHeight]{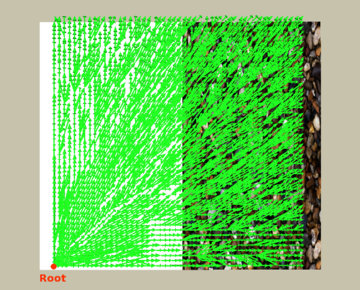}}
        {\caption{\enspace \small Standard RRT* - 2500 steps} \label{fig:tree_construction_c}}

      \end{subfloatrow}
    }
     \CommonHeightRow
    {
      \begin{subfloatrow}[3]
        \ffigbox[\FBwidth]
        {\includegraphics[height=\CommonHeight]{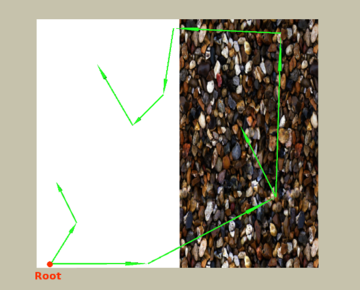}}
        {\caption{\enspace \small Perception-aware RRT* - 10 steps} \label{fig:tree_construction_d}}
        \ffigbox[\FBwidth]
        {\includegraphics[height=\CommonHeight]{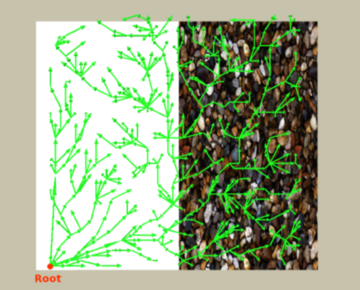}}
        {\caption{\enspace \small Perception-aware RRT* - 500 steps} \label{fig:tree_construction_e}}
        \ffigbox[\FBwidth]
        {\includegraphics[height=\CommonHeight]{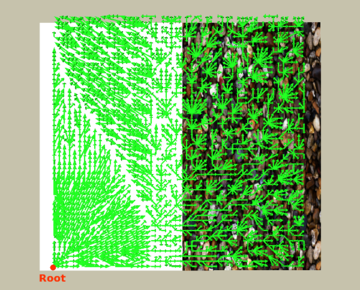}}
        {\caption{\enspace \small Perception-aware RRT* - 2500 steps} \label{fig:tree_construction_f}}

      \end{subfloatrow}
    }

    \caption{\small Evolution of the optimal policy tree after different iterations. From left to right, we plot the state of the 
      tree respectively after 10, 500 and 2500 sampling steps. 
      In \ref{fig:tree_construction_a}-\ref{fig:tree_construction_c} the planner follows the standard RRT* strategy, \ie, the shortest path, without taking into account the information from the vision sensor. 
      By contrast, our framework \ref{fig:tree_construction_d}-\ref{fig:tree_construction_f} computes trajectories that attempt to minimize the pose uncertainty using the photometric information gain. \vspace{-1.5em}
}
    \label{fig:tree_construction}
  }
\end{figure*}
  
  At a high level, the rapidly-exploring random trees algorithm explores the state space to compute the optimal path $\mathcal{T}$ from the start location to each point in the space.
  In particular, the tree is composed of a set of vertices $V$ representing elements of the state space along with their associated pose covariances.
  Each vertex $v \in V$ has a list of neighboring vertices $v.N$, a state $v.x$, where $x \in SE(3)$, a state covariance $v.\TSigma$, a cost value $v.c$,
  a unique parent vertex $v.p$, and the photometric information gain $v.\Information$ associated to the camera viewpoint at $v.x$.
  Figure \ref{fig:tree_representation} depicts the properties of the tree.
  The graph is incrementally built by sampling new states and connecting them to the existing vertices, propagating the covariances towards
  the new one. Furthermore, since each location $x$ is associated with a view and a depth map, we can anticipate
  what the robot will see in a specific position and compute the associated photometric information gain. 
  The algorithm makes use of the dense image-to-model alignment strategy,
  presented in Section \ref{information_gain}, to compute the predicted information gain and update the pose covariance accordingly. 

  Each nominal trajectory $\mathcal{T}_i \in \mathcal{P}$ is described by a sequence of $N_i$ waypoints $v_{j}^{i}$, where each of them is a vertex of the tree.
  To solve the problem of finding the plan that represents the best trade-off between path length and pose estimation accuracy, we propose a cost function that weighs both the distance between waypoints,
  and the pose covariances.
  Among all the candidate paths $\mathcal{P}$, we select the trajectory $\mathcal{T}_i \in \mathcal{P}$ that minimizes the following function:
  
   \begin{equation}
     J(\mathcal{T}_i) = \sum_{j=1}^{N_i} \alpha \text{ Dist}(v_{j}^{i}.x, v_{j-1}^{i}.x) + (1-\alpha) \Tr(v_{j}^{i}.\Sigma) \label{cost_function} 
   \end{equation}  
  where $\alpha$ is the trade\textendash off factor between path length minimization and information maximization, and $\text{Dist}(\cdot, \cdot)$ computes the distance between the two locations.
  It should be noticed that, by choosing to minimize the sum of the trace of all the pose covariances, we suggest the algorithm to seek the trajectory that keeps 
  small the camera pose uncertainty along the candidate path.
  We choose the trace to include the visual information into the cost function following the considerations in \cite{haner2011optimal}. In particular, minimizing the trace of the pose covariance matrix
(A-optimality) guarantees that the majority of the state space dimensions is considered (in contrast to the D-optimality), but does not require us to compute all the eigenvalues (E-optimality).
  
  Algorithm \ref{rrt} describes the proposed Perception-aware RRT*.
At each iteration, the algorithm samples a new state from the state space, then it creates and adds the associated vertex to the tree. After that, 
the vertices near the new one are selected through the function $\text{Near}()$. This function looks for the vertices whose states are within a ball of radius $\rho$, 
defined as follows (see \cite{karaman2011sampling}):

 \begin{equation}
    \begin{aligned}
      \rho \propto \left( \frac{\log(n)}{n} \right )^{\frac{1}{d}}.
    \end{aligned}
  \end{equation}  
In the above equation, the radius depends on the dimension of the state $d$ and on the number of state vertices $n$.
It is important to notice that, before checking for adjacent vertices, the function $\text{Nearest}()$ selects the nearest node without checking if it is inside the ball of radius $\rho$.
This is required especially during the first iterations, when the tree is very sparse and, thus, the $\text{Near}()$ function can easily return an empty list.
The new vertex is then connected along a minimum cost path to one of the neighbors (lines 10-23). 
In particular, for each element in the neighborhood we first check whether there is a safe connection between the two vertices, \ie, whether there are any collisions along the path. 
The collision radius $c$ (see Algorithm \ref{rrt}) depends on the geometrical structure of the robot and is provided as an input parameter.
Afterwards, the pose uncertainty associated with the current $v_{\text{near}}$ vertex is propagated using (\ref{covariance_propagation}) and updated according to the photometric information gain expected 
from receiving an image measure when reaching the state $x_{\text{new}}$. 
Finally, we check whether the overall cost of connecting $v_{\text{near}}$ to $v_{\text{new}}$ (which represents the cost of the candidate path $\mathcal{T}$ through those waypoints) 
is smaller than the current minimum, and update it if necessary.

In the final stage of the algorithm, we update the tree connections following the strategy proposed in \cite{karaman2011sampling}: the vertices in the neighborhood are 
visited, updating their parent relationships in the tree if the path through $v_{\text{new}}$ is more convenient. This procedure is referred as $\text{RewireTree}()$.
During the $\text{RewireTree}()$ procedure we iterate through the subtrees of each $v_{\text{near}}$ whose parent relationships has been changed with $v_{\text{new}}$ to propagate the updated 
covariances and maintain the child nodes consistent after rewiring.

The output of the overall procedure is a connected tree, from which we can extract the optimal policy to a generic goal vertex following the parent relationships from
the final to the start state. Figure \ref{fig:tree_construction} shows the evolution of the tree at different iteration steps and compares the standard RRT* with our perception-aware formulation.

\subsection{Online Perception-aware Planning}
\label{online-perception-rrbt}

\begin{figure}[t!]
  \centering
  \ffigbox{}
  {
    \CommonHeightRow
    {
      \begin{subfloatrow}[3]
        \ffigbox[\FBwidth]
        {\includegraphics[height=\CommonHeight]{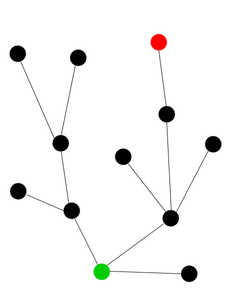}}
        {\caption{}\label{fig:update_tree_a}}
        \ffigbox[\FBwidth]
        {\includegraphics[height=\CommonHeight]{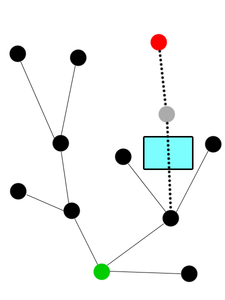}}
        {\caption{}\label{fig:update_tree_b}}
        \ffigbox[\FBwidth]
        {\includegraphics[height=\CommonHeight]{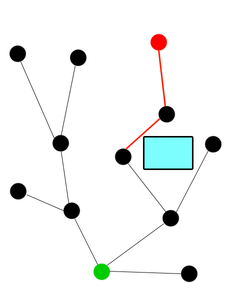}}
        {\caption{}\label{fig:update_tree_c}}
      \end{subfloatrow}
    }
    \CommonHeightRow
    {
      \begin{subfloatrow}[3]
        \ffigbox[\FBwidth]
        {\includegraphics[height=\CommonHeight]{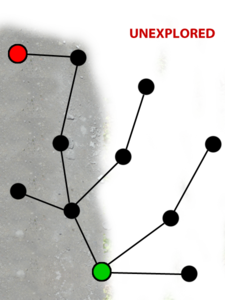}}
        {\caption{}\label{fig:update_tree_d}}
        \ffigbox[\FBwidth]
        {\includegraphics[height=\CommonHeight]{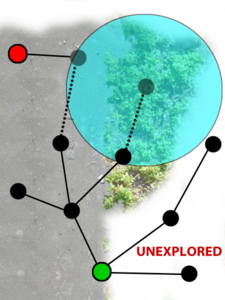}}
        {\caption{}\label{fig:update_tree_e}}
        \ffigbox[\FBwidth]
        {\includegraphics[height=\CommonHeight]{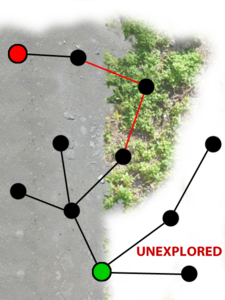}}
        {\caption{}\label{fig:update_tree_f}}
      \end{subfloatrow}
    }

    \caption{\small Online update steps during exploration: Figures \subref{fig:update_tree_a}-\subref{fig:update_tree_c} depict the subtree invalidation and rewiring update when an
      obstacle is spotted, while \subref{fig:update_tree_d}-\subref{fig:update_tree_f} show how the tree is rewired when new
      photometric information is available from the scene. \vspace{-1.5em}}
    \label{fig:update_tree}
  }
\end{figure}

Given an initially optimal path, we can now start exploring the environment.
When new parts of the scene are revealed, the current trajectory might become non-optimal or even infeasible in case of obstacles.
One possibility would be to recompute the tree from scratch after every map update but this would be costly and computationally intractable to have the system integrated into an MAV application.
For this reason, we propose to update the planning tree \emph{on-the-fly} by only processing vertices and edges affected by new information.
This online update is illustrated in Figure~\ref{fig:update_tree} and its fundamental steps are depicted in Algorithm~\ref{rrt_online}.

\begin{algorithm}[t]
  \caption{\enspace Online perception-aware RRT*}
  \begin{small}
    \begin{algorithmic}
      \State
      01: \textbf{while} $1$ \textbf{do} \State
      02: \ \ $\text{UpdateCollisionMap}()$ \State
      03: \ \ $\text{UpdatePhotometricInformationMap}()$ \State 
      04: \ \ $V_{\text{colliding}} = \text{NewCollidingVertices}()$ \State
      05: \ \ $\text{InvalidateSubTree}(V_{\text{colliding}})$ \State 
      06: \ \ Run $\text{PerceptionAwareRRT*}$ \ref{rrt} \State
      07: \ \ $V_{\text{inf}} = \text{UpdatedVertices}()$ \State
      08: \ \ \textbf{for all} $v_{\text{inf}} \in V_{\text{inf}} \textbf{do}$ \State
      09: \ \ \ \  $\Information_v$ =  $\Information_{v}^{new}$ \State 
      10: \ \ \ \ $\text{RewireTree}()$ \State
      11: \ \ \textbf{end for} \State
      12: \textbf{end while} \State
    \end{algorithmic}
  \end{small}
  \label{rrt_online}
\end{algorithm}

Consider an initial planning tree as in Figure~\ref{fig:update_tree_a}, that is grown from a starting point (indicated by a green circle) to a desired end point location (the red circle).
Whenever a new obstacle is spotted, the respective edge and the affected subtree get invalidated and regrown (lines 04-06) as in Figure~\ref{fig:update_tree_b}. 
Note that the \texttt{SampleUnexplored()} function is 
now bounded within the subspace corresponding to the invalidated subtree, which results in a drastically reduced number of iterations compared to fully regrowing the RRT* tree from scratch.
The second scenario in Figures~\ref{fig:update_tree_d} to~\ref{fig:update_tree_f} demonstrates the case of gaining areas with distinctive photometric information. 
As newly discovered areas provide photometric information, as shown in Figure~\ref{fig:update_tree_e}, the neighboring vertices are updated 
by the \texttt{RewireTree()} procedure (lines 07-10 in Algorithm~\ref{rrt_online}). Potentially better connections are considered to form a new path with lower costs (Figure~\ref{fig:update_tree_f}).

\section{Experiments}\label{experiments_sec}

To validate the proposed method, we run experiments assuming both known and unknown scenarios. 
The formers (Section \ref{sec:experiments_known_sec}) aim to to show how, in contrast to standard strategies, our perception-aware path planner selects trajectories that favor highly-textured areas.
In the latter ones (Section \ref{sec:online_exp}) we demonstrate the capability to adapt the perception-aware plan in an online fashion as new information is available from the environment. 
Furthermore, we test our approach within a complete visual navigation system that explores, localizes itself and computes trajectory considering the visual information from the scene.

\subsection{Experiments in Known Scenarios}\label{sec:experiments_known_sec}

We evaluate the approach in both simulated and real scenes.
In the simulated experiments, we used Blender to generate photorealistic, textured scenes and render images with the associated depth maps.
We assume a down-looking camera in both simulated and real scenarios.
In contrast to the experiments in the following sections, here we assume to have full knowledge about the map and the texture in the scene.

Our framework can handle 6DoF state representations (\ie, $(x,y,z,\rho,\phi,\theta)$).
However, since we assume flight in near-hover conditions, without loss of generality, we can omit the roll and pitch angles (\ie, $\rho = 0, \phi = 0$).
Furthermore, since the orientation angle $\theta$ does not affect the information-gain computation with down-looking camera, we can also omit $\theta$ 
(in the experiments in unknown scenarios, described in Section \ref{sec:online_exp}, we consider also the front-looking configuration, \ie we plan including the yaw angle).
\\
\subsubsection{Simulation Results}

\begin{figure}[t!]
  \centering
  \ffigbox{}
  {
    \CommonHeightRow
    {
      \begin{subfloatrow}[3]
        \ffigbox[\FBwidth]
        {\includegraphics[height=\CommonHeight]{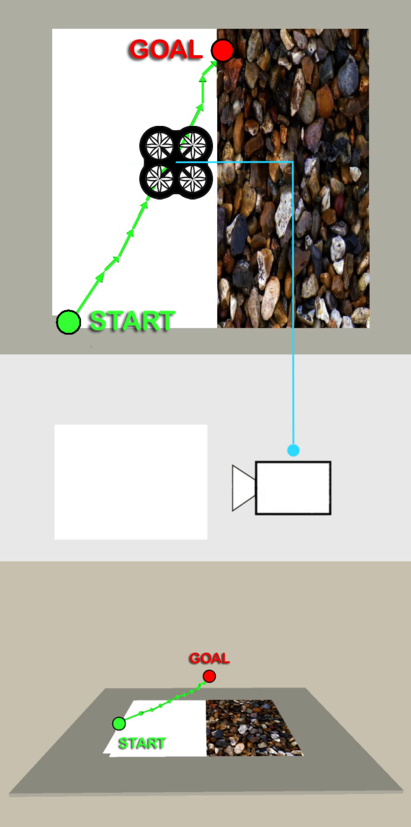}}
        {\caption{\enspace $\alpha = 0.9$}\label{fig:simple_plan_a}}
        \ffigbox[\FBwidth]
        {\includegraphics[height=\CommonHeight]{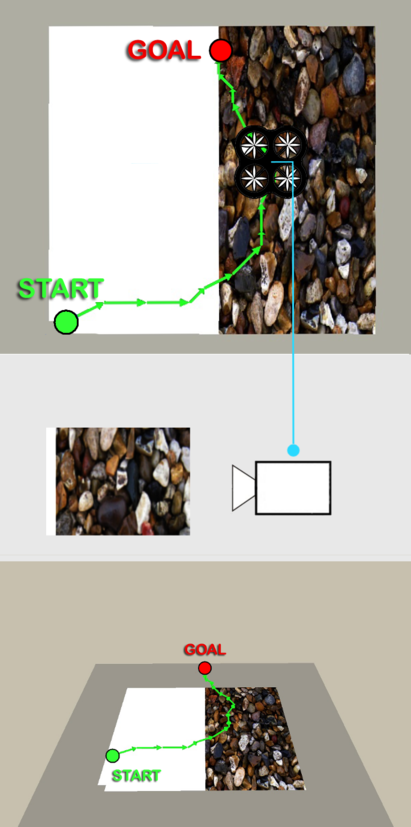}}
        {\caption{\enspace $\alpha = 0.1$} \label{fig:simple_plan_b}}
        \ffigbox[\FBwidth]
        {\includegraphics[height=\CommonHeight]{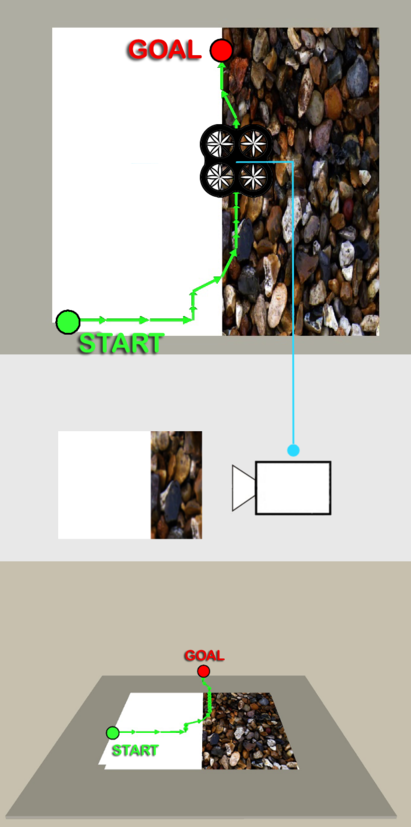}}
        {\caption{\enspace $\alpha = 0.4$} \label{fig:simple_plan_c}}

      \end{subfloatrow}
    }

    \caption{\small Results of the experiment with two textures. The images are extracted from the 
		      graphical interface of the planner where the measures are displayed as a colored point cloud. 
		      The green arrows indicate the optimal path. 
		      The experiments with $\alpha = $0.9, 0.1 and 0.4 are shown from left to right.
		      The first row shows the top view above the scene and in the second one we depict the image from the down-looking camera acquired at an intermediate pose along the trajectory. 
		      In the third row a 3D perspective view is depicted. \vspace{-1.5em}
}
    \label{fig:simple_plan}
  }
\end{figure}

\begin{figure*}[th!]
  \centering
  \ffigbox{}
  {
    \CommonHeightRow
    {
      \begin{subfloatrow}[4]
        \ffigbox[\FBwidth]
        {\includegraphics[height=\CommonHeight]{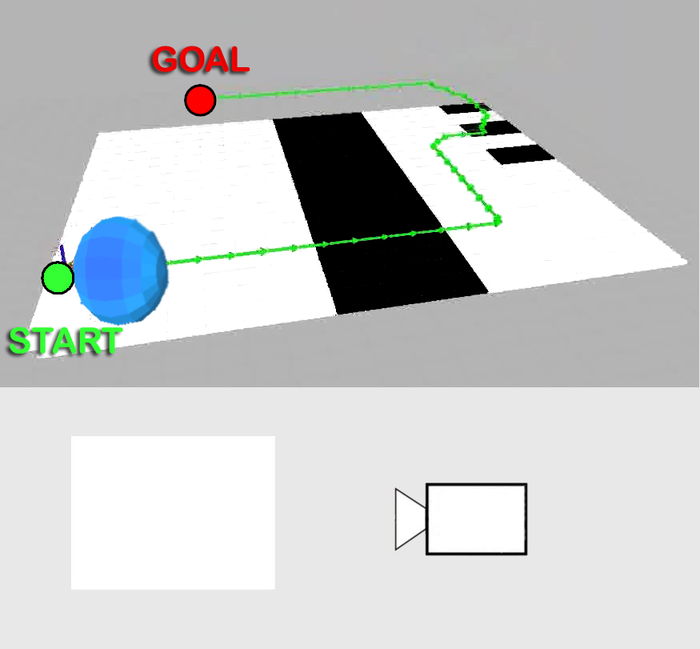}}
        {\caption{}\label{fig:texture_plan_cov_a}}
        \ffigbox[\FBwidth]
        {\includegraphics[height=\CommonHeight]{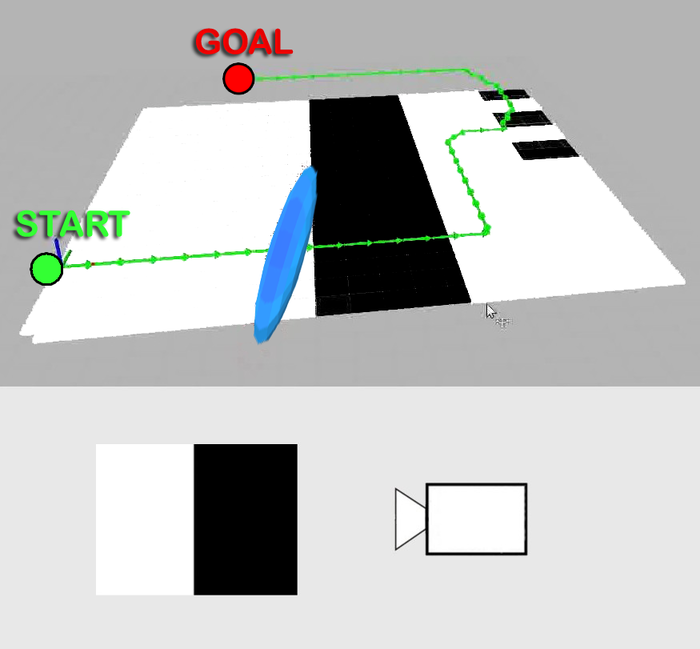}}
        {\caption{} \label{fig:texture_plan_cov_b}}
        \ffigbox[\FBwidth]
        {\includegraphics[height=\CommonHeight]{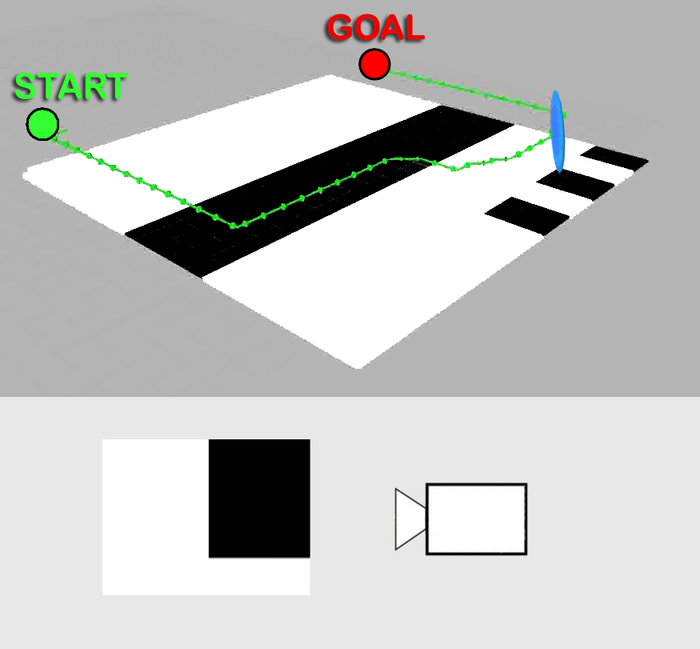}}
        {\caption{} \label{fig:texture_plan_cov_c}}
        \ffigbox[\FBwidth]
        {\includegraphics[height=\CommonHeight]{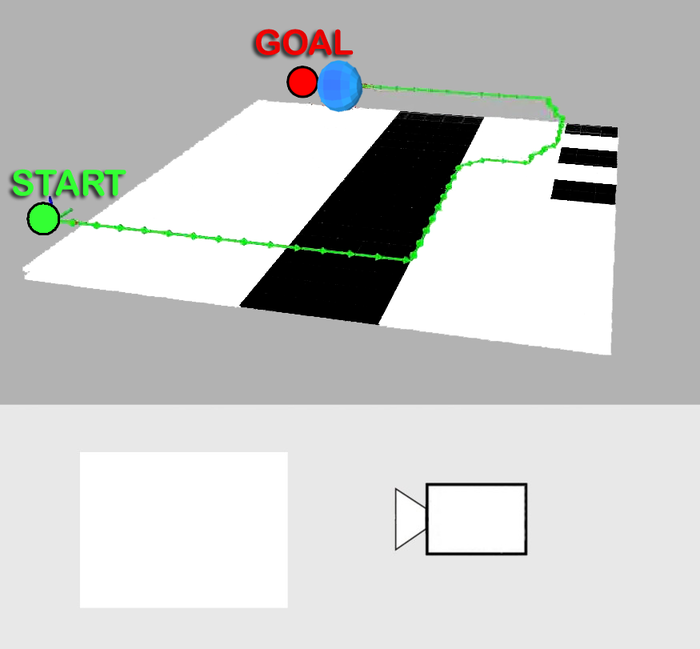}}
        {\caption{} \label{fig:texture_plan_cov_d}}

      \end{subfloatrow}
    }

    \caption{\small Uncertainty propagation samples from the computed optimal policy. In this test $\alpha$ is set to $0.1$ In the first row the red cloud indicates the 
      covariance at the given position, while in the second row the camera image rendered in that position is displayed. In the second and in the third column
      is it possible to see how the uncertainty is reduced first along the x axis and then along the y and the z axes. \vspace{-1.5em}
}
    \label{fig:texture_plan_cov}
  }
\end{figure*}

We set up two different simulation scenarios to prove that our approach can effectively compute the optimal trajectory with respect to the uncertainty reduction. 
In particular, we discuss the effect of the trade-off factor $\alpha$ \eqref{cost_function} on the computed path. 
In the first experiment (Figure \ref{fig:simple_plan}), the scene is divided into two areas: the first one textureless and the second one with texture. 
The second scenario (Figure \ref{fig:texture_plan_cov}) contains texture that only reduces the uncertainty along one dimension, \eg, with zero intensity gradient along specific directions. 
In particular, the scene is characterized by black and white stripes along the $x$ and $y$ directions.
This test is designed to demonstrate how our approach predicts the pose uncertainty specifically for each state dimensions and plans accordingly.

For each simulated scenario, we render images at different locations.
This way, we can compute the photometric information gain with different camera viewpoints and update the predicted pose covariance along the trajectory.

In the first test (Figure \ref{fig:simple_plan}), the space is limited to a $10 \times 10$ meter area. 
The states $(x,y,z) = (0.0,0.0,2.0)$ and $(x,y,z) = (2.0,9.0,2.0)$ are chosen respectively as the start and the goal state. 
We split the scenario in two areas: the first is textureless, while the second one is highly-textured. 
For this experiment we keep the camera height at 2 meters above the ground.
As the start and the goal state are both located in the white zone, selecting a straight trajectory that only minimizes the distance leads to a viewpoint sequence without texture.

We run three tests setting the parameter $\alpha$ to 0.9, 0.1 and 0.4, respectively.
In the first one, the planner penalizes long paths, while in the second one a higher cost is associated to trajectories with high uncertainty.
Finally, in the last one, the computed trajectory is a trade-off between localization accuracy and trajectory length.

Figure \ref{fig:simple_plan} shows that in the case $\alpha = 0.9$, the planner correctly selects the trajectory close to the shortest one (\ie, a line). 
In the second case $\alpha = 0.1$, the optimal viewpoint sequence includes the textured area, to keep the uncertainty small as long as possible along the path. 
Finally, in the case $\alpha = 0.4$ the computed path keeps the pose covariances small, but, since more weight is given to the distance term in the cost function,
the planner reduces the trajectory length as much as possible.

Within the second simulation (see Figure \ref{fig:texture_plan_cov}), we demonstrate how the proposed approach seeks to maximize the information gain along all dimensions of the space domain. 
As explained in section \ref{perception-rrbt}, we can achieve this behavior through the proposed cost function, which tries to minimize the sum of the traces of the pose covariances.

In this case the space is constrained to a $20 \times 20$ meter area. 
The start state is still set at $(x,y,z) = (0.0,0.0,2.0)$ and the final state is $(x,y,z) = (5.0,19.0,2.0)$. 
The simulated scenario is composed of three types of texture: one completely white and the remaining two with black and white stripes along the $y$ and the $x$ axes respectively. 
Furthermore, we set $\alpha = 0.1$, \ie, we look for the path that minimizes the uncertainty.
Figure \ref{fig:texture_plan_cov} shows the resulting optimal path. 
The planner suggests first reaching the area with the stripe along the $y$ axis, then navigating to the stripe along the $x$ axis to minimize the uncertainty along both directions. 
Furthermore, as shown in Figure~\ref{fig:texture_plan_cov}, we can also reduce the uncertainty of the $z$ dimension. 
When two gradients with known relative position are available we can gain information about the depth.

In Table \ref{tab:results1} we also report the comparison between the trajectory length and the trace of the pose covariance matrix for different values of $\alpha$.
In particular, we run tests in the two scenarios varying $\alpha$ from 0.05 to 0.95 with a 0.1 step.
We perform 10 runs per tests averaging the trajectory length, the mean trace along the path, and the trace at the goal location.
In the first test, as we give more importance to the pose uncertainty minimization, the length of the trajectories varies between 9.21 $m$ and 12.91 $m$, while the mean and the goal state traces are reduced.
The results vary almost linearly because the optimal trajectory changes smoothly between a straight towards the goal (shorter paths) and two almost orthogonal segments (safer paths), as we change the value of $\alpha$. 
Conversely, in the second experiment we can observe three different behaviors. When more importance is given to the trajectory length, the planner selects a straight path towards the goal (over the 
area with no texture), thus, the uncertainty is very high. On the other hand, with small $\alpha$ values, the trajectories selected are similar to the one depicted in Figure \ref{fig:texture_plan_cov}.
However, when $\alpha$ is between 0.35 and 0.65, the planner computes paths that reach only the first black stripe, without going over the black squares on the other side of the space. In this way, 
it is not possible to reduce the uncertainty with respect to the $y$ axis, but the trajectory is shorter. 

\begin{figure}[b!]
  \centering
  \ffigbox{}
  {
    \CommonHeightRow
    {
      \begin{subfloatrow}[2]
        \ffigbox[\FBwidth]
        {\includegraphics[height=\CommonHeight]{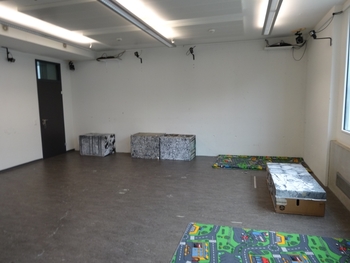}}
        {\caption{} \label{fig:real_exp_setup_a}}
        %
        %
        \ffigbox[\FBwidth]
        {\includegraphics[height=\CommonHeight]{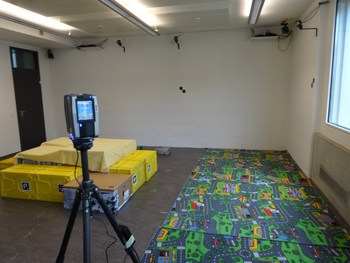}}
        {\caption{} \label{fig:real_exp_setup_c}}
      \end{subfloatrow}
    }

    \caption{\small Two different scenario setups in our laboratory.}
    \label{fig:real_exp_setup}
  }
\end{figure}

\begin{table*}[]
\caption{\small Comparison between the trajectory length and the pose uncertainty for different $\alpha$ values.}
\label{tab:results1}
\begin{center}
\resizebox{1.0\linewidth}{!} {
\begin{tabular}{|*{1}{c|}|*{6}{c|}}
\hline
{} & \multicolumn{3}{|c|}{First Scenario (Figure \ref{fig:simple_plan})} & \multicolumn{3}{|c|}{Second Scenario (Figure \ref{fig:texture_plan_cov})} \\
\cline{2-7}
$\alpha$ & Avg. Length [m] & Avg. Mean Trace & Avg. Goal Trace & Avg. Length [m] & Avg. Mean Trace & Avg. Goal Trace \\
\hline
0.05 & 12.91 & 2.1 & 1.0 & 40.12 & 7.60 & 9.05 \\
\hline
0.15 & 12.91 & 3.56 & 1.5 & 40.45 & 7.75 & 9.25 \\
\hline
0.25 & 11.50 & 6.4 & 1.8 & 40.23 & 7.23 & 9.28 \\
\hline
0.35 & 11.05 & 9.05 & 4.0 & 23.36 & 35.23 & 18.35 \\
\hline
0.45 & 11.05 & 9.12 & 3.87 & 23.33 & 37.11 & 17.98 \\
\hline
0.55 & 11.04 & 10.36 & 4.23 & 23.45 & 36.59 & 19.12 \\
\hline
0.65 & 10.5 & 25.4 & 8.34& 23.5 & 38.67 & 18.81 \\
\hline
0.75 & 9.89 & 30.22 & 18.45 & 19.67 & 65.61 & 76.34 \\
\hline
0.85 & 9.67 & 30.67 & 18.12 & 19.63 & 67.04 & 78.24 \\
\hline
0.95 & 9.21 & 30.5 & 19.09 & 19.64 & 69.12 & 79.67 \\
\hline

\end{tabular}
}
\end{center}
\end{table*}

\subsubsection{Real Experiments}

While simulated scenarios are well-suited to demonstrate the capabilities of the proposed framework, a real-world experimental setup is important to prove the effectiveness of the approach in actual environments.
The 3D surface model of the scene was computed using a Faro 3D laser scanner\footnote{http://www.faro.com} to gather a fine-grained point cloud representation of the scenario. 
After recording the scans, we generated the state space and computed trajectories given different start and goal points.
In addition, we used a quadrotor with a down-looking camera to perform the computed trajectories.

\begin{figure*}[h!t]
  \centering
  \ffigbox{}
  {
    \CommonHeightRow
    {
      \begin{subfloatrow}[3]
        \ffigbox[\FBwidth]
        {\includegraphics[height=4.0cm]{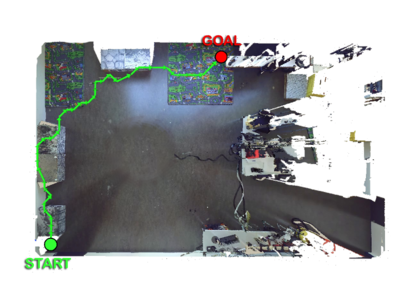}}
        {}
        \ffigbox[\FBwidth]
        {\includegraphics[height=4.0cm]{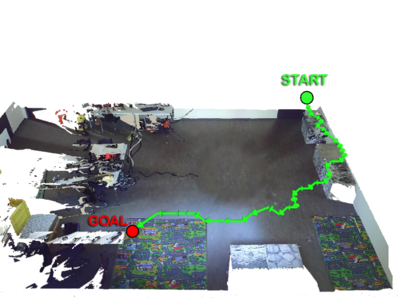}}
        {\caption{\enspace$\alpha = 0.1$} \label{fig:real_exp11_a}}
        \ffigbox[\FBwidth]
        {\includegraphics[height=4.0cm]{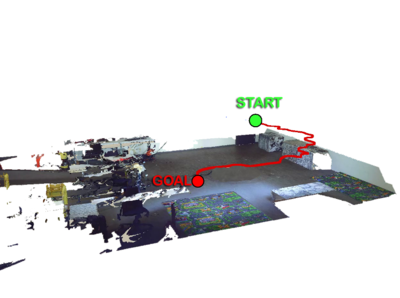}}
        {}
      \end{subfloatrow}
    }
    \CommonHeightRow
    {
      \begin{subfloatrow}[3]
        \ffigbox[\FBwidth]
        {\includegraphics[height=4.0cm]{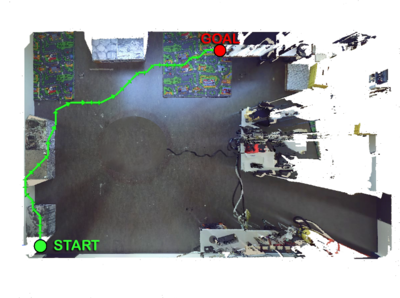}}
        {}
        \ffigbox[\FBwidth]
        {\includegraphics[height=4.0cm]{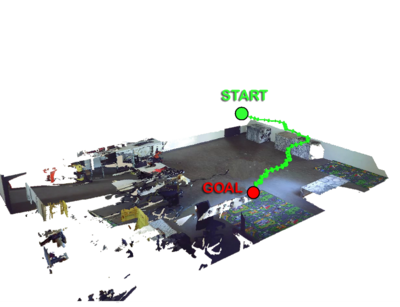}}
        {\caption{\enspace$\alpha = 0.3$} \label{fig:real_exp11_b}}
        \ffigbox[\FBwidth]
        {\includegraphics[height=4.0cm]{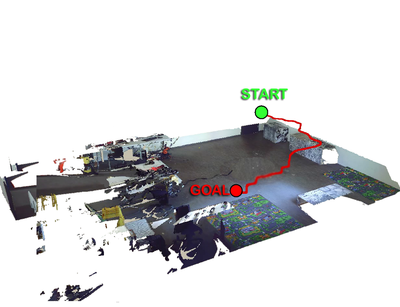}}
        {}
      \end{subfloatrow}
    }
    \CommonHeightRow
    {
      \begin{subfloatrow}[3]
        \ffigbox[\FBwidth]
        {\includegraphics[height=4.0cm]{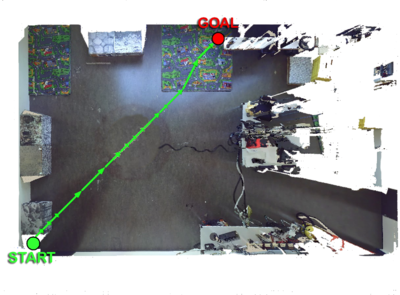}}
        {}
        \ffigbox[\FBwidth]
        {\includegraphics[height=4.0cm]{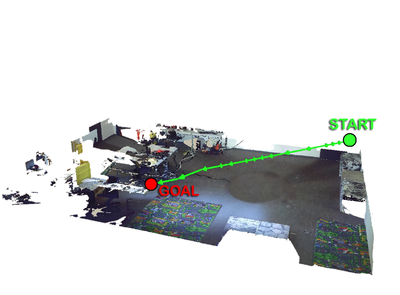}}
        {\caption{\enspace$\alpha = 1.0$} \label{fig:real_exp11_c}}
        \ffigbox[\FBwidth]
        {\includegraphics[height=4.0cm]{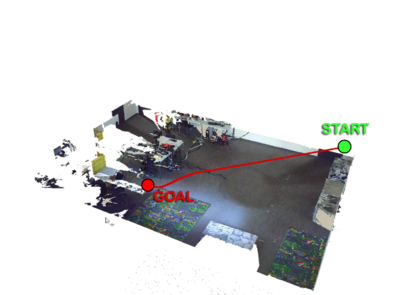}}
        {}
      \end{subfloatrow}
    }

    \caption{\small Results of the first scenario tests in a real environment. Each row shows the computed optimal path for each planner parametrization. 
      In particular \ref{fig:real_exp11_a} and \ref{fig:real_exp11_b} depict the computed trajectories with $\alpha = 0.1$ and $\alpha = 0.3$, while \ref{fig:real_exp11_c} displays the standard RRT* output ($\alpha = 1.0$).
      The first two columns from the left show two different perspective views of each trajectory in the scenario, while in the rightmost column depicts the interpolated trajectory in red.}
    \label{fig:real_exp11}
  }
\end{figure*}

\begin{figure*}[h!t]
  \centering
  \ffigbox{}
  {
    \CommonHeightRow
    {
      \begin{subfloatrow}[3]
        \ffigbox[\FBwidth]
        {\includegraphics[height=3.0cm]{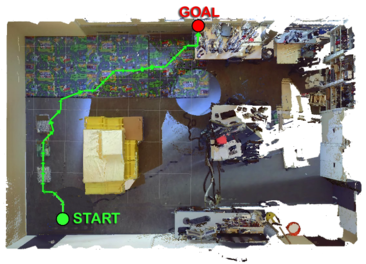}}
        {}
        \ffigbox[\FBwidth]
        {\includegraphics[height=3.0cm]{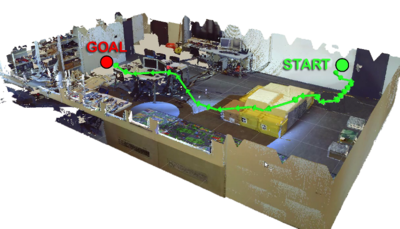}}
        {\caption{\enspace Perception-aware RRT*} \label{fig:real_exp3_a}}
        \ffigbox[\FBwidth]
        {\includegraphics[height=3.0cm]{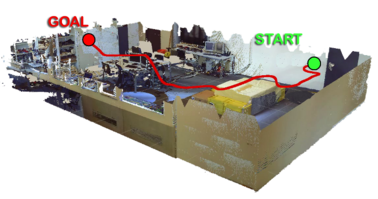}}
        {}
      \end{subfloatrow}
    }
    \CommonHeightRow
    {
      \begin{subfloatrow}[3]
        \ffigbox[\FBwidth]
        {\includegraphics[height=3.0cm]{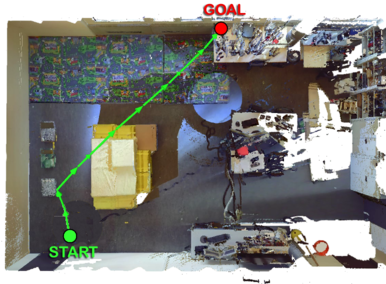}}
        {}
        \ffigbox[\FBwidth]
        {\includegraphics[height=3.0cm]{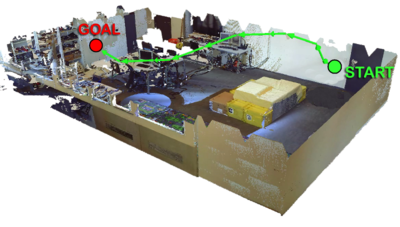}}
        {\caption{\enspace Standard RRT*} \label{fig:real_exp3_b}}
        \ffigbox[\FBwidth]
        {\includegraphics[height=3.0cm]{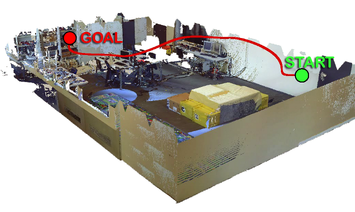}}
        {}
      \end{subfloatrow}
    }

    \caption{\small The trajectories computed in the second scenario experiment. Our Perception-aware RRT* with $\alpha = 0.1$ \ref{fig:real_exp3_a} is compared with the standard RRT* \ref{fig:real_exp3_b}.}
    \label{fig:real_exp3}
  }
\end{figure*}

We set up two scenarios to test our approach with different texture and object arrangements. For each scenario, a full 3D scan of the room was acquired.
Figure \ref{fig:real_exp_setup} shows different scenario setups during scan acquisitions.

%

In the first configuration, shown in Figure \ref{fig:real_exp_setup_a}, the scene is left without texture on the floor, apart from the area near the walls, where we added highly-textured boxes and 
carpets (Figure \ref{fig:inf_gain_b} shows the photometric information gain at different locations in the scene).
The start position is set close to the room door, while the goal position is located in the opposite corner of the room. 
We compare the standard RRT* planner ($\alpha = 1.0$) with our Perception-aware RRT* using two different configurations:
in the first one, more importance is given to pose-uncertainty minimization ($\alpha = 0.1$); in the second one, the planner is asked to 
select the trajectory that also favors short path lengths ($\alpha = 0.3$).
As shown in Figure \ref{fig:real_exp11}, while the obvious solution for the standard RRT* planner is to go straight to the goal along the diagonal of the room (see Figure \ref{fig:real_exp11_c}), 
our framework understands that it is not the best path with respect to visual localization, as no texture for sufficient pose estimation is available.
In particular, when $\alpha = 0.1$, it selects the trajectory along the walls, retrieving photometric information from the scenario. 
It should also be noticed that setting $\alpha = 0.3$ results in a path that keeps the robot close to the walls to minimize the pose covariance while at the same time reducing the path length as much as possible.
Thus, compared to the $\alpha = 0.1$ experiment, the trajectory is shorter but with a less accurate pose estimation. 

The last scenario, shown in Figure \ref{fig:real_exp_setup_c}, was set with some boxes with uniform color in the center of the room and with texture-rich 
carpets and boxes along the walls. 
The start and the goal states are the same as in previous scenarios.
Our Perception-aware planner with $\alpha = 0.1$ is again compared with the standard RRT* approach. 
The minimum-length trajectory (\ie, the output of the RRT* planner) is obtained selecting viewpoints over the boxes in the middle of the room (see Figure \ref{fig:real_exp3_b}).
However, this path is poor in photometric information, thus, the strategy implemented by our approach chooses a trajectory along the walls, circumventing the central boxes and keeping the 
quadrotor height low to maximize the uncertainty reduction (cf. Figure \ref{fig:real_exp3_a}).
Inspecting the results in Figure \ref{fig:real_exp3}, it can be seen that our planner increases the height in some parts of the trajectory. 
Although, in general, higher depth values reduce the photometric information for the translational components, higher waypoints have larger scene coverage.
In this scenario, as some parts of the scene are poor in texture, photometric information increases with height because this boosts the possibility of acquiring richer texture from other areas. 

\begin{figure}[h!]
  \centering
  \ffigbox{}
  {
    \CommonHeightRow
    {
      \begin{subfloatrow}[2]
        \ffigbox[\FBwidth]
        {\includegraphics[height=\CommonHeight]{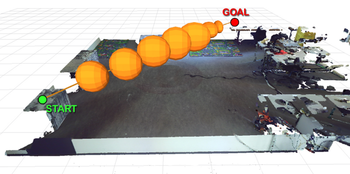}$\rule{0.7px}{55px}$}
        {}
        \ffigbox[\FBwidth]
        {\includegraphics[height=\CommonHeight]{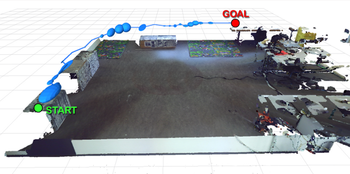}}
        {}
      \end{subfloatrow}
    }
     \CommonHeightRow
    {
      \begin{subfloatrow}[2]
        \ffigbox[\FBwidth]
        {\includegraphics[height=\CommonHeight]{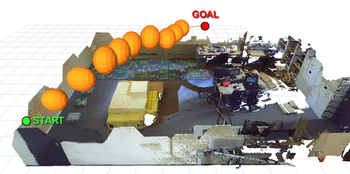}\hspace{0.96em}$\rule{0.7px}{55px}$}
        {\caption{\enspace Standard RRT*} \label{fig:real_exp12_a}}
        \ffigbox[\FBwidth]
        {\includegraphics[height=\CommonHeight]{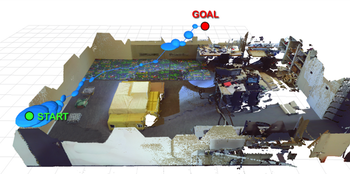}}
        {\caption{\enspace Perception-aware RRT*} \label{fig:real_exp12_b}}
      \end{subfloatrow}
    }

    \caption{\small Pose estimation uncertainties plotted for each experiment. The figure compares the standard planner output \ref{fig:real_exp12_a} with the proposed perception-aware results \ref{fig:real_exp12_b}.
      The covariances and the sequences of viewpoints computed with the standard RRT* are pictured in orange, while the perception-aware RRT* is depicted in blue.
      From top to bottom, the figure shows the computed trajectories in the two scenarios. }
    \label{fig:real_exp12}
  }
\end{figure}

\begin{figure}[]
  \centering
  
        {\includegraphics[height=3.6cm]{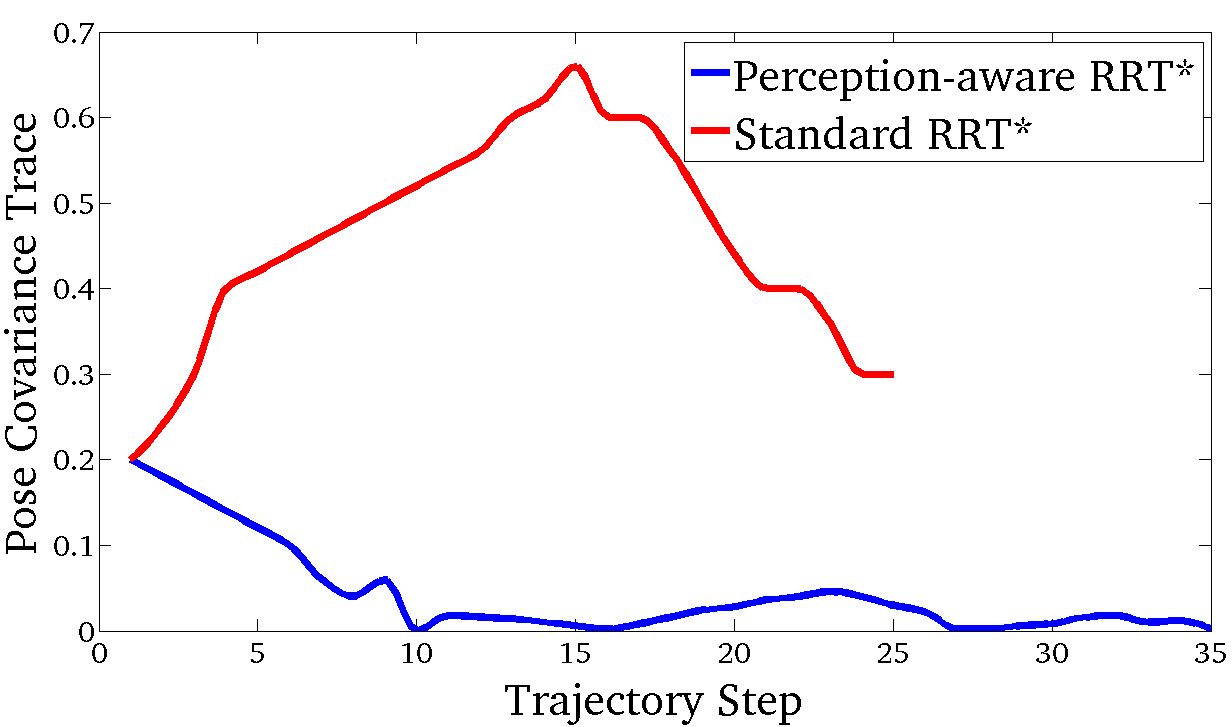}}\\
        %
        {\includegraphics[height=3.6cm]{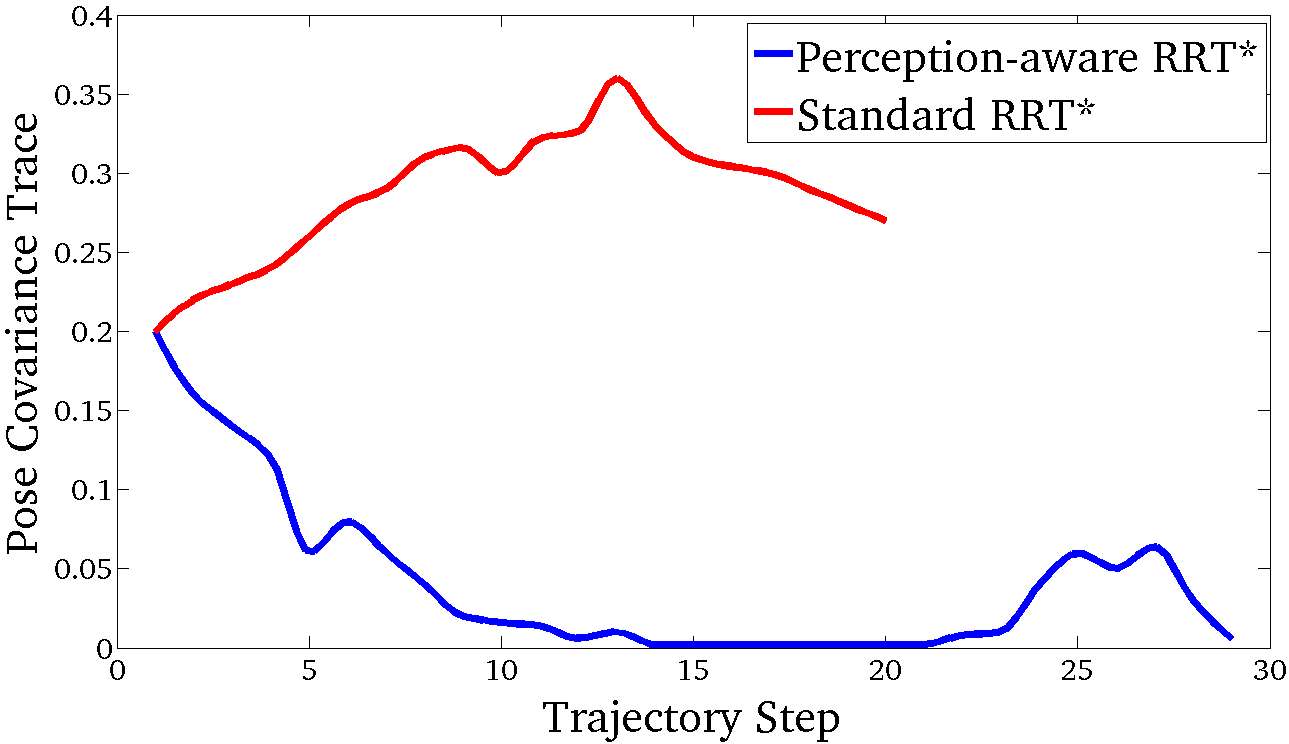}}

    \caption{\small Comparison of the pose covariance estimates along the trajectories computed with the standard RRT* and our perception-aware RRT*.
    The plot at the top depicts the comparison of the pose covariance trace for the first scenario (see Figure \ref{fig:real_exp_setup_a}), while the bottom one shows the results of the experiments 
    on the second scenario (see Figure \ref{fig:real_exp_setup_c}).
      Despite the Standard RRT* trajectories are shorter, the pose covariance uncertainty along the paths is significantly higher than our perception-aware RRT*.}
    \label{fig:real_exp_plot}
  
\end{figure}

Finally, in Figures \ref{fig:real_exp12} and \ref{fig:real_exp_plot} we compare the pose covariance estimates relative to the trajectories computed with the standard RRT* and the proposed Perception\textendash aware RRT*. 
The plots clearly show that we can effectively reduce the pose uncertainty by selecting paths over highly-textured areas. 
Conversely, since the standard RRT* planner does not take into account any photometric information, the resulting trajectories provide a small amount of texture to
the visual localization system and, thus, they are characterized by larger covariance values.

\subsection{Experiments in Unknown Scenarios}
\label{sec:online_exp}

In this section we discuss the experiments in unknown scenarios. We first describe the architecture of the visual navigation system that performs online localization, map reconstruction and planning.
Afterwards, we present the results achieved in both simulated and real environments.
\\
\subsubsection{System Overview}
\label{sec:system-overview}

\begin{figure}[t]
  \begin{center}
    \includegraphics[width=0.99\columnwidth]{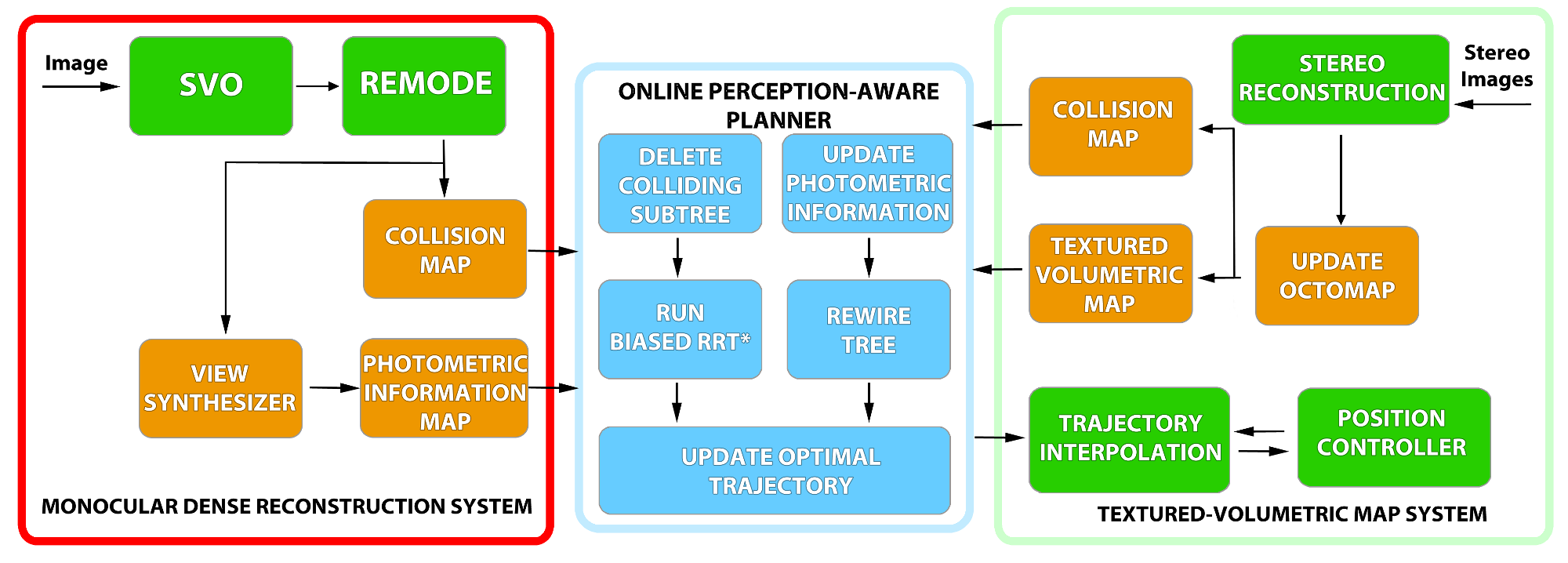}
    \caption{\small Block diagram of the online perception-aware planning system. \vspace{-1.5em}}
    \label{fig:sys_overview}
  \end{center}
\end{figure}

We consider an MAV that explores an unknown environment by relying only on its camera to perform localization, dense scene reconstruction and optimal trajectory planning.
We have integrated the online perception-aware planner with two different mapping systems (see Figure \ref{fig:sys_overview}): a monocular dense reconstruction system that generates a 
point cloud map, and a volumetric system that uses stereo camera input. 

In the monocular system, the localization of the quadrotor runs onboard, providing the egomotion estimation to perform navigation and stabilization.
To achieve real-time performance, the dense map reconstruction and the online perception-aware path planning runs off-board on an Intel i7 laptop with a GPU, in real-time.

At each time step $k$, the quadrotor receives a new image 
to perform egomotion estimation. 
We use the Semi-direct monocular Visual Odometry (SVO) proposed in \cite{forster2014svo}, which allows us to estimate the quadrotor motion in real-time.
The computed pose $\T_{k,w}$ and the relative image are then fed into the dense map reconstruction module (REMODE \cite{Pizzoli2014}, a probabilistic, pixelwise depth estimator to compute dense depthmaps).
Afterwards, the dense map provided by the reconstruction module is sent to the path planning module and is used to update both the collision map (using Octomap ~\cite{hornung2013octomap}) 
and the photometric information map. 
The last one is then used to update $\Information_v$ for each vertex affected by the map update.
Finally, we update the optimal trajectory following the procedure described in Algorithm \ref{rrt_online}.

For the textured volumetric map system, we take input from a stereo camera, perform egomotion estimation with SVO as above, and compute a dense depth map with OpenCV's Block Matcher.
The estimated camera pose from SVO and the point cloud produced from the depth map are used to update a textured OctoMap.
This volumetric map serves as a collision map, when it is queried for occupancy, and is used to synthesize views and compute photometric information gain during planning, when it is queried for texture.
This pipeline runs in real time onboard an MAV's embedded single board computer (an Odroid XU3 Lite) using a map with $5 cm$ resolution, and with the input images downsampled by a factor of $4$ to $188\times120$, and throttled down to $1Hz$.
However, we evaluate this system in simulation, and for the experiments in Sec. \ref{sec:sim_experiments}, we run the simulation, visual pipeline, planner, and control software all on a laptop with an Intel i7 processor.
\\
\subsubsection{Real Experiments}
\label{sec:online_exp_real}

\begin{figure}[th!]
  \centering
  \ffigbox{}
  {
    \CommonHeightRow
    {
      \begin{subfloatrow}[3]
        \ffigbox[\FBwidth]
        {\includegraphics[height=\CommonHeight]{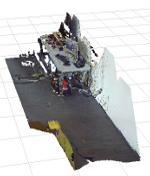}}
        {}
        \ffigbox[\FBwidth]
        {\includegraphics[height=\CommonHeight]{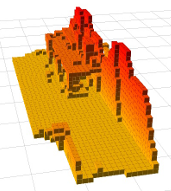}}
        {\caption{}\label{fig:inf_gain_map_a}}
        \ffigbox[\FBwidth]
        {\includegraphics[height=\CommonHeight]{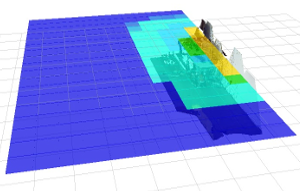}}
        {}
      \end{subfloatrow}
    }
    \CommonHeightRow
    {
      \begin{subfloatrow}[3]
        \ffigbox[\FBwidth]
        {\includegraphics[height=\CommonHeight]{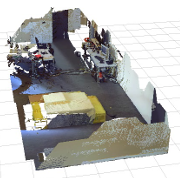}}
        {}
        \ffigbox[\FBwidth]
        {\includegraphics[height=\CommonHeight]{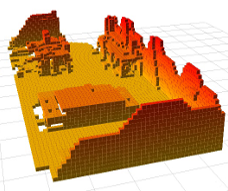}}
        {\caption{}\label{fig:inf_gain_map_b}}
        \ffigbox[\FBwidth]
        {\includegraphics[height=\CommonHeight]{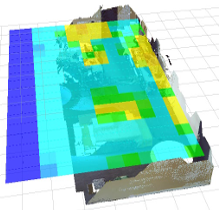}}
        {}
      \end{subfloatrow}
    }
    \CommonHeightRow
    {
      \begin{subfloatrow}[3]
        \ffigbox[\FBwidth]
        {\includegraphics[height=\CommonHeight]{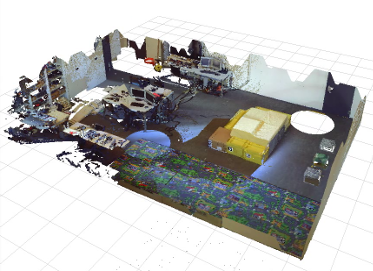}}
        {}
        \ffigbox[\FBwidth]
        {\includegraphics[height=\CommonHeight]{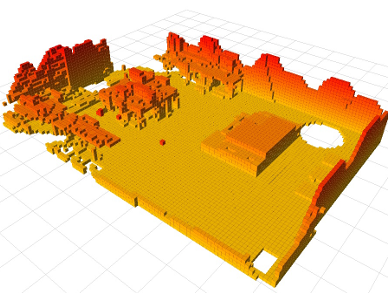}}
        {\caption{}\label{fig:inf_gain_map_c}}
        \ffigbox[\FBwidth]
        {\includegraphics[height=\CommonHeight]{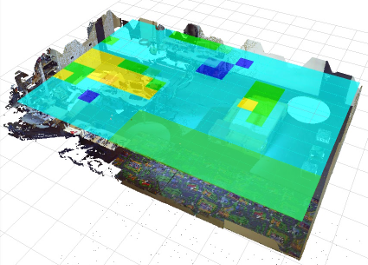}}
        {}
      \end{subfloatrow}
    }

    \caption{\small Three different exploration stages of a scene (rows). The
      first column shows the scene layout, the second column the collision map
      and the third the computed photometric information gain. \vspace{-1.5em}
      }
    \label{fig:inf_gain_map}
  }
\end{figure}

Before presenting the experimental results, 
we motivate our approach by discussing how the photometric information distribution changes over time when exploring an unknown environment.
Figure~\ref{fig:inf_gain_map} shows the map for collision avoidance and the photometric information gain at different exploration stages. 
In the photometric information map, warm (yellowish) colors refer to camera viewpoints exhibiting a higher amount of texture, while the cool (bluish) ones indicate less informative areas.
In Figure~\ref{fig:inf_gain_map_a} the almost unexplored scene has very little valuable information to compute a reliable trajectory.
Hence, standard planners, that calculate trajectories only once (without performing online updates), compute sub-optimal trajectories or even collide with undiscovered objects.
Therefore, an online approach is needed to integrate the information from newly unexplored areas and re-plan accordingly. 
While exploring, the collision map and photometric information get updated (see Figures~\ref{fig:inf_gain_map_b} and~\ref{fig:inf_gain_map_c}) 
and become useful to update the optimal trajectory.

For the experiment in unknown real scenarios, we set up three scenarios with different object and carpet arrangements to vary the texture and the 3D structure of the scene.
In the first scenario, the camera on the MAV is downward-looking, while in the last one choose a front-looking configuration with an angle of 45 degrees with respect to the ground plane.
We made experiments with two different camera setups to investigate the influence of the camera viewpoint on the optimal trajectory computation. 
Intuitively, the front-looking configuration provides more information since also areas far from the quadrotor are observed. 
Conversely, with the downward-looking configuration, the pose estimation algorithm is more reliable, but less information is captured from the scene.
Finally, in all the experiments we set $\alpha = 0.1$ to increase the importance of the pose uncertainty minimization.

\begin{figure*}[th]
\centering
  \ffigbox{}
  {
    \CommonHeightRow
    {
      \begin{subfloatrow}[3]
        \ffigbox[\FBwidth]
        {\includegraphics[height=3.1cm]{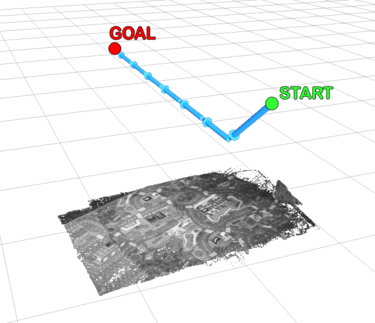}}
        {\caption{}\label{fig:exp_eval_a}}
        \ffigbox[\FBwidth]
        {\includegraphics[height=3.1cm]{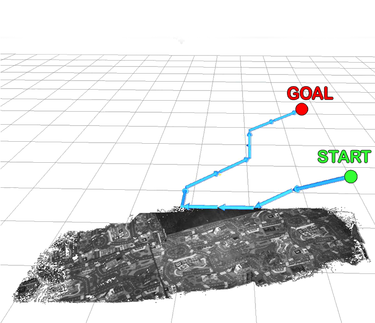}}
        {\caption{}\label{fig:exp_eval_b}}
        \ffigbox[\FBwidth]
        {\includegraphics[height=3.1cm]{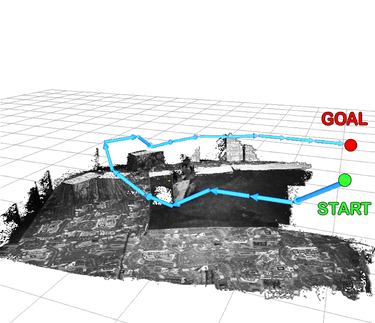}}
        {\caption{}\label{fig:exp_eval_c}}
      \end{subfloatrow}
    }
    \CommonHeightRow
    {
      \begin{subfloatrow}[3]
        \ffigbox[\FBwidth]
        {\includegraphics[height=3.1cm]{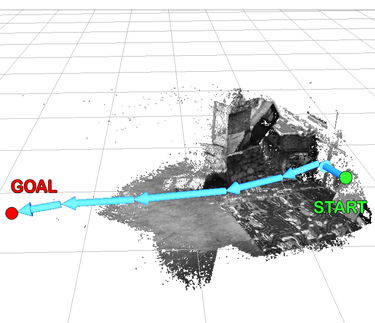}}
        {\caption{}\label{fig:exp_eval_d}}
        \ffigbox[\FBwidth]
        {\includegraphics[height=3.1cm]{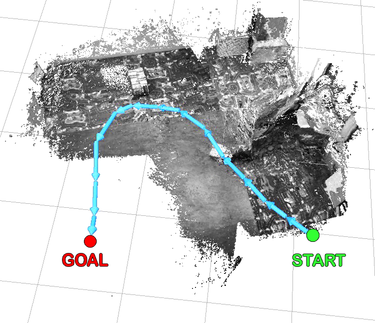}}
        {\caption{}\label{fig:exp_eval_e}}
        \ffigbox[\FBwidth]
        {\includegraphics[height=3.1cm]{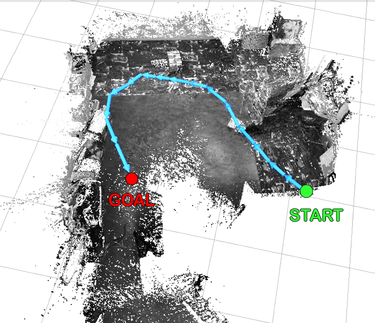}}
        {\caption{}\label{fig:exp_eval_f}}
      \end{subfloatrow}
    }
    \CommonHeightRow
    {
      \begin{subfloatrow}[3]
        \ffigbox[\FBwidth]
        {\includegraphics[height=3.1cm]{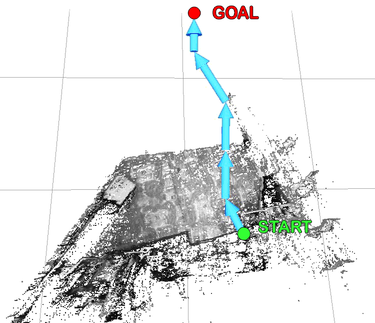}}
        {\caption{}\label{fig:exp_eval_g}}
        \ffigbox[\FBwidth]
        {\includegraphics[height=3.1cm]{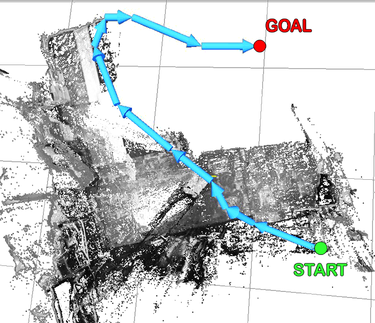}}
        {\caption{}\label{fig:exp_eval_h}}
        \ffigbox[\FBwidth]
        {\includegraphics[height=3.1cm]{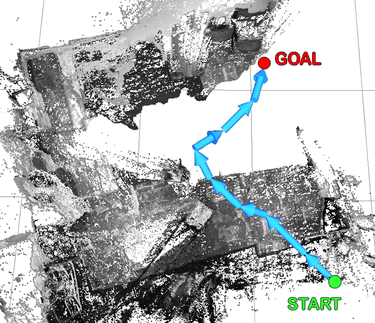}}
        {\caption{}\label{fig:exp_eval_i}}
      \end{subfloatrow}
    }

      \caption{\small Experimental results in three real scenarios (rows). The first column shows the initially computed trajectories, only having little information of the environment available. 
      The second and third column demonstrate the update of the trajectory as new information is gathered by updating the scene. \vspace{-1.5em}
      }
        \label{fig:exp_eval}
      }
\end{figure*}

In all the scenarios, we put highly-textured carpets along the walls, while the floor in the center of the room is left without texture (\ie, with a uniform color).
We also place some boxes on the carpets and near the walls. In the first scenario, we also put an obstacle in the center of the room. 
At the beginning of the exploration, the planner shows a behavior similar in all the experiments (see Figures \ref{fig:exp_eval_a}, \ref{fig:exp_eval_d} and \ref{fig:exp_eval_g}).
The information about the scene is very low, thus, our approach computes a simple straight trajectory to the goal.
As the robot explores the environment, the trajectory is updated by preferring areas with high photometric information.
In the first scenario, we can observe that, since a new obstacle (a box near the center of the room) is spotted at the end of the exploration, 
the previous trajectory (cf. Figure \ref{fig:exp_eval_b}) is invalid 
and a new collision-free one is computed (see Figure \ref{fig:exp_eval_c}). However, to guarantee the availability of photometric information, 
our approach correctly suggests to fly over the textured boxes and not toward the center of the room.

A front-looking camera configuration (second and third scenario) provides photometric information about areas distant from the current MAV pose. 
As a consequence, we can obtain an optimal trajectory, with respect to pose uncertainty minimization, earlier with respect to the previous experiment 
(see Figures \ref{fig:exp_eval_e}, \ref{fig:exp_eval_f}, \ref{fig:exp_eval_h} and \ref{fig:exp_eval_i}).
In the final stage of the exploration of the third scenario, the obstacles near the goal are spotted (see Figure \ref{fig:exp_eval_i}).
As a consequence, the trajectory in Figure \ref{fig:exp_eval_h} is invalidated. Despite more texture are available, 
flying over the top left corner of the room is not anymore convenient due to the presence of the boxes near the goal position. Therefore, our approach correctly updates the trajectory.
In this last experiment, we can also observe that, even if the reconstructed map is noisy, 
our approach correctly computes the best trajectory with respect to the pose uncertainty minimization. 
Sparse methods would be more affected by the reconstruction error compared to the used dense image-to-model alignment strategy which can effectively capture the photometric information. 
\\
\subsubsection{Simulated Experiments}
\label{sec:sim_experiments}

\begin{figure*}[t]
  \centering
  \ffigbox{}
  {
    \CommonHeightRow
    {
    
    \begin{subfloatrow}[3]
         \ffigbox[\FBwidth]
        {\includegraphics[height=3.5cm]{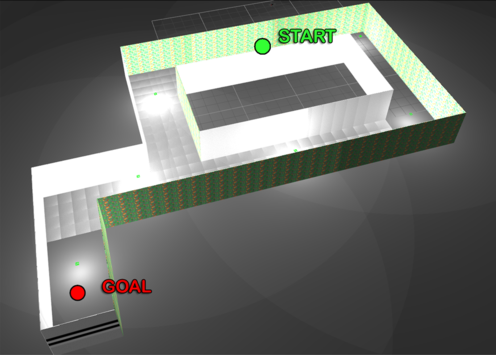}}
        {\caption{}\label{fig:test_world}}
        \ffigbox[\FBwidth]
        {\includegraphics[height=3.5cm]{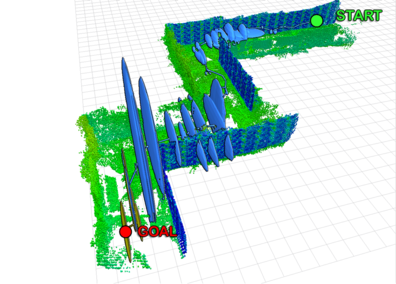}}
        {\caption{\enspace RRT*}\label{fig:test_world_rrt_3}}
        \ffigbox[\FBwidth]
        {\includegraphics[height=3.5cm]{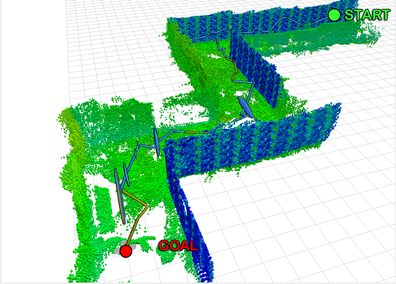}}
        {\caption{\enspace Perception-aware}\label{fig:test_world_perc_3}}
      \end{subfloatrow}
    
      \begin{subfloatrow}[3]
        \ffigbox[\FBwidth]
        {\includegraphics[height=3.5cm]{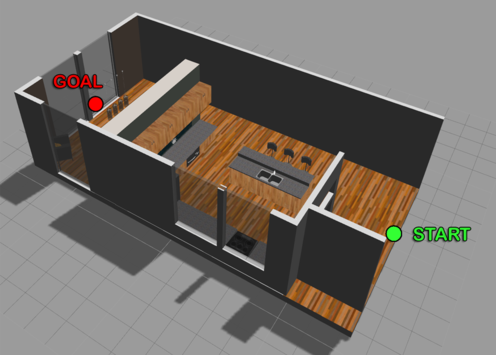}}
        {\caption{}\label{fig:kitchen_world}}
        \ffigbox[\FBwidth]
        {\includegraphics[height=3.5cm]{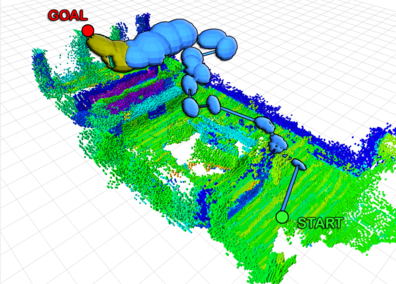}}
        {\caption{\enspace RRT*}\label{fig:kitchen_world_rrt_3}}
        \ffigbox[\FBwidth]
        {\includegraphics[height=3.5cm]{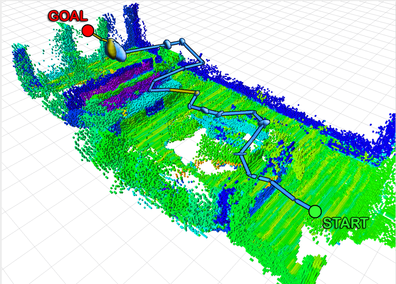}}
        {\caption{\enspace Perception-aware}\label{fig:kitchen_world_perc_3}}
      \end{subfloatrow}
      
    }
    \caption{\small Exploration trial in the \emph{labyrinth} \subref{fig:test_world} and in the \emph{kitchen} \subref{fig:kitchen_world} simulated environments.
      The trajectories computed by the RRT* planner are shown in Fig. \subref{fig:test_world_rrt_3} for the labyrinth scenario and in 
      Fig. \subref{fig:kitchen_world_rrt_3} for the kitchen,
      while the ones computed with the perception aware planner are shown in \subref{fig:test_world_perc_3} and in \subref{fig:kitchen_world_perc_3}, respectively.
      The Textured OctoMaps are visualized with a color corresponding to the mean intensity over all of the observed faces, with red representing high intensity, and purple representing low intensity. 
      The pose covariance at each waypoint is shown as an ellipse, with the most recent update in orange, and the rest of the trajectory in blue. 
}
    \label{fig:simulated_experiment}
  }
\end{figure*}

To further evaluate the performance of our system in wider and more complex scenarios, we also run tests in a simulated environment, using the components described in Sec. \ref{sec:system-overview}.
Two trials were performed in environments simulated with \emph{Gazebo}, one designed to explicitly test perception (\emph{labyrinth}) and one designed to simulate a real world environment (\emph{kitchen}).
The labyrinth scenario is designed with flat and highly-textured walls to test the capability of our perception-aware planner to choose the MAV orientations that maximize the amount of photometric information.
The quadrotor starts in one of the two long corridors in the scene (see \ref{fig:test_world}) and is asked to reach the goal location that is located at $25m$ from the start location.
In the kitchen world (see \ref{fig:kitchen_world}), the MAV begins at a position that is separated by two walls from the goal location, which is $12.5m$ away.
We compare the performance of the standard RRT* planner and our perception-aware planner in Figs. \ref{fig:simulated_experiment} and \ref{fig:quantitative_results}.

\subsubsection{Discussion}
\label{sec:discussion}
\begin{figure*}[t]
  \centering
  \ffigbox{}
  {
    \CommonHeightRow
    {
      \begin{subfloatrow}[3]
        \hspace*{-0.3cm}
        \ffigbox[\FBwidth]
        {\includegraphics[height=4.0cm]{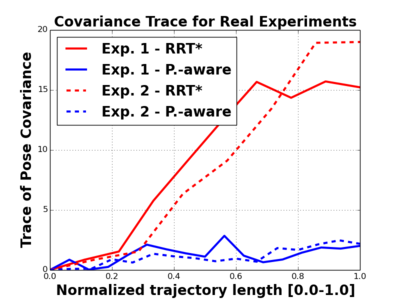}}
        {\caption{}\label{fig:real_exp}}
        \ffigbox[\FBwidth]
        {\includegraphics[height=4.0cm]{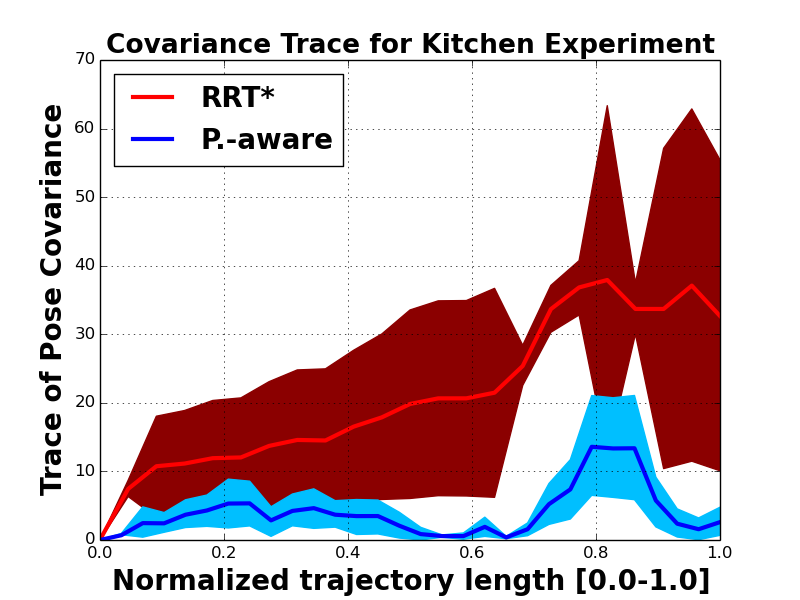}}
        {\caption{}\label{fig:kitchen_exp}}
        \ffigbox[\FBwidth]
        {\includegraphics[height=4.0cm]{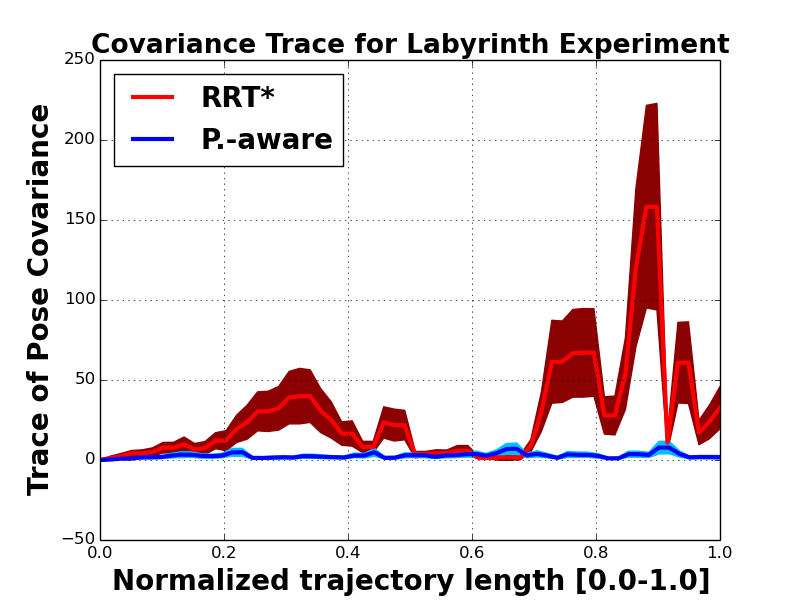}}
        {\caption{}\label{fig:labyrinth_exp}}

      \end{subfloatrow}
      
    }
    \caption{\small Quantitative results for our experiments showing the evolution of the MAV's pose covariance during the planned trajectory. Fig. \subref{fig:real_exp} shows results of the real world experiments. 
    Figs. \subref{fig:kitchen_exp} and \subref{fig:labyrinth_exp} show the simulated {\it kitchen} and {\it labyrinth} trials, respectively.  
    The sequence of viewpoints for each trial result in different length trajectories, so the length of each one is normalized to one.
    For each simulated experiment, we conducted 15 trials, normalized the trajectories, and inferred Gaussian distributions at each point in a set of equally-spaced samples along a normalized trajectory.  
    In \subref{fig:kitchen_exp} and \subref{fig:labyrinth_exp}, each solid line represents the mean over all of the trials, and the colored band is the 95\% confidence interval. \vspace{-1.5em}
}
    \label{fig:quantitative_results}
  }
\end{figure*}

The qualitative results shown for the real world (Fig. \ref{fig:exp_eval}) and simulated (Fig. \ref{fig:simulated_experiment}) experiments show that the perception-aware planner does indeed choose 
trajectories that allow the MAV to observe more photometric information.
Quantitatively, this results in a dramatic improvement in the uncertainty of the vehicle's pose estimate.
The results in Fig. \ref{fig:quantitative_results} show that the pose uncertainty, measured as the trace of the covariance matrix and visualized as ellipses in Fig. \ref{fig:simulated_experiment}, 
is up to an order of magnitude smaller when the planner considers the texture of the environment.

In both of the simulated experiments, the RRT* and perception aware planners both reached the goal location in all trials.
On average, for the \emph{labyrinth} it took $718s$ and $715s$, respectively, and for \emph{kitchen} it took $578s$ and $580s$, respectively.
The results are shown in Figs. (\ref{fig:test_world_rrt_3}) and \ref{fig:test_world_perc_3} for the labyrinth tests and in Figs. \ref{fig:kitchen_world_rrt_3} and \ref{fig:kitchen_world_perc_3} for the kitchen ones.
The most important distinction in this performance comparison is the pose uncertainty across the trajectory.
The two planners produce similar trajectories in terms of waypoint positions, but the covariances for the RRT* trajectory are much larger due to the desired yaw angles that are chosen for the waypoints.
The proposed perception aware planner specifically optimizes the waypoint position and yaw angle (i.e. \emph{where to look}) in order to minimize this pose uncertainty.
As a consequence, the trajectory computed with our strategy has low pose uncertainty values, while the RRT* trajectory, which does not consider the visual information, leads to very low localization
accuracy, which can make the navigation infeasible due to the high risk of collisions.

\section{Conclusion and Future Work}\label{conclusion_sec}

In this paper, we gave a new double twist to the problem of planning under uncertainty by proposing a framework 
(called Perception-aware Path Planning) to incorporate the photometric information of a scene, in addition to geometric one, to compute trajectories with minimum 
localization uncertainty of vision-control robots in goal-reaching tasks.

To avoid the caveats of feature-based localization systems (i.e., dependence of feature type and use-defined thresholds), 
we proposed to use \emph{dense, direct methods} to compute the Fisher information matrix directly from the intensity values of every pixel in the image. 
We used Lie-Group-based propagation to approximate the localization uncertainty up to the fourth order. 
Finally, we proposed to adapt trajectories in an online fashion, considering also scenarios with no prior knowledge about the map.

The proposed framework is general and can easily be adapted to different robotic platforms and scenarios. 
As an application, we showed how the proposed framework can be adapted to the well known RRT* planner.

The proposed framework was validated in both real and simulated environments.
Finally, we presented the integration and demonstration of the overall system into a real quadrocopter performing vision-based localization, dense map reconstruction, and online perception-aware planning. 
The results clearly show that our framework can generate trajectories that outperforms standard path-planning approaches in terms of vision-based localization accuracy.

We believe that this will translate into safer trajectories for vision-controlled robots.
Future work will investigate solutions to predict the photometric information gain in unexplored areas using past knowledge. 
This way, we will be able to reach better estimates of the optimal trajectory even before discovering all the scene elements.
Finally, we plan to include dynamic constraints and control effort in the optimization process to generate smoother trajectories.
\begin{small}
\bibliographystyle{IEEEtran}
\bibliography{bibliography}
\end{small}

\end{document}